\documentclass[1p]{elsarticle}

\usepackage{amsmath}
\usepackage{amssymb}
\usepackage{multirow}
\usepackage{float}
\usepackage{graphicx}
\usepackage{cite}
\usepackage{xcolor}
\usepackage[ruled,vlined]{algorithm2e}
\usepackage{hyperref}
\usepackage{natbib}
\usepackage{subcaption}

\journal{Information Sciences}
    
\SetKwInput{kwParams}{Params}

\begin{document}

\begin{frontmatter}
\title{AttrE2vec: Unsupervised Attributed Edge Representation Learning}
\author[PWR]{Piotr Bielak}
\author[PWR]{Tomasz Kajdanowicz}
\author[PWR,UND]{Nitesh V. Chawla}

\address[PWR]{Department of Computational Intelligence, Wroclaw University of Science and Technology, Poland}
\address[UND]{Department of Computer Science and Engineering, University of Notre Dame, Notre Dame, IN, USA}

\begin{abstract}
Representation learning has overcome the often arduous and manual featurization of networks through (unsupervised) feature learning as it results in embeddings that can apply to a variety of downstream learning tasks. The focus of representation learning on graphs has focused mainly on shallow (node-centric) or deep (graph-based) learning approaches. While there have been approaches that work on homogeneous and heterogeneous networks with multi-typed nodes and edges, there is a gap in learning edge representations. This paper proposes a novel unsupervised inductive method called \texttt{AttrE2Vec}, which learns a low-dimensional vector representation for edges in attributed networks. It systematically captures the topological proximity, attributes affinity, and feature similarity of edges.
Contrary to current advances in edge embedding research, our proposal extends the body of methods providing representations for edges, capturing graph attributes in an inductive and unsupervised manner. Experimental results show that, compared to contemporary approaches, our method builds more powerful edge vector representations, reflected by higher quality measures (AUC, accuracy) in downstream tasks as edge classification and edge clustering. It is also confirmed by analyzing low-dimensional embedding projections. 
\end{abstract}


\end{frontmatter}

\section{Introduction}

Complex networks, included attributed and heterogeneous networks, are ubiquitous --- from recommender systems to citation networks and biological systems~\citep{Hua}. These networks present a multitude of machine learning problem statements, including node classification, link prediction, and community detection. A fundamental aspect of any such machine learning (ML) task, transductive or inductive, is the availability of featurized data. Traditionally, researchers have identified several network characteristics suited to specific ML tasks and used them for the learning algorithm. This practice is arduous as it often entails customizing to each specific ML task, and also is limited to the computable characteristics. 

This has led to a surge in (unsupervised) algorithms and methods that learn embeddings from the networks, such that these embeddings form the featurized representation of the network for the ML tasks~\citep{Zhang2018a, Wu2019, Lia, Chami2020, autoweight}. This area of research is generally notated as representation learning in networks. Generally, these embeddings generated by representation learning methods are agnostic to the end use-case, as they are generated in an unsupervised fashion. Traditionally, the focus was on representation learning on homogeneous networks, i.e. the networks that have singular type of nodes and edges, and also do not have attributes attached to the nodes and edges~\citep{Lia}. 

\begin{figure}
    \centering
    \includegraphics[width=0.9\columnwidth]{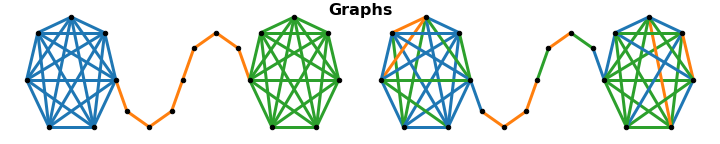}
    \includegraphics[width=0.7\columnwidth]{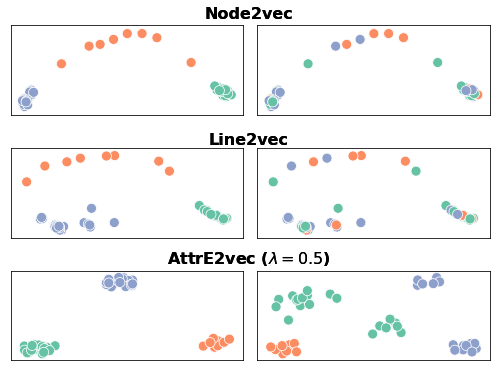}
    \caption{Our proposed AttrE2vec model compared to other methods in the task of an attributed graph embedding. Colors denote edge features. On the left we can see a graph, where the features are aligned to substructures of the graph. On the right, the features were shuffled (ca. 50\%). Traditional approaches fail to build robust representations, whereas our method includes features information to construct the embedding vectors.\label{fig:intro-figure}}
\end{figure}

Existing representation learning models mainly focus on transductive learning, where a model can only be trained using the entire input graph. It means that the model requires all the nodes and a fixed structure of the network in the training phase, e.g., Node2vec \citep{Grover2016}, DeepWalk \citep{Perozzia} and GCN \citep{Kipf2019}, to some extent. Besides, there have been methods focused on heterogeneous networks that incorporate different typed nodes and edges in a network, as well as content at each node~\citep{DongMetapath2vec:Networks, covid19gcn}. 

On the other hand, a less explored and exploited approach is the inductive setting. In this approach, only a part of the network is used to train the model to infer embeddings for new nodes. Several attempts have been made in the inductive setting including EP-B \citep{Garcia-Duran2017}, GraphSAGE \citep{Hamiltona}, GAT \citep{Velickovic2018}, SDNE \citep{Wanga}, TADW \citep{Yang2015}, AHNG\citep{ahng} or PVECB \citep{Lan2020ImprovingContent}. There is also recent progress on heterogeneous graph embedding, e.g., MIFHNE \citep{Li2020Multi-sourceEmbedding} or models based on graph neural networks~\citep{ZhangHeterogeneousNetwork}.

State-of-the-art network embedding techniques are mostly unsupervised, i.e., aim at learning low-dimensional representations that preserve the structure of an input graph, e.g., GraphSAGE \citep{Hamiltona}, DANE \citep{Gao2018a}, line2vec \citep{Bandyopadhyay}, RCAN \citep{Chen2020RelationEmbedding}. Nevertheless, semi-supervised or supervised methods can learn vector representations but for a specific downstream prediction task, e.g., TADW \citep{Yang2015} or FSCNMF \citep{Bandyopadhyay2018}. Hence it has been shown in the literature that not much supervision is required to learn the embeddings.

%Attributes
In recent years, proposed models mainly focus on the graphs that do not contain attributes related to nodes and edges \citep{Lia}. It is especially noticeable for edge attributes. The majority of proposed approaches consider node attributes only, omitting the richness of edge feature space while learning the representation. Nevertheless, there have been successfully introduced such models as DANE \citep{Gao2018a}, GraphSAGE \citep{Hamiltona}, SDNE \citep{Wanga} or CAGE \citep{Nozza2020CAGE:Embedding} which make use of node features and EGNN \citep{Kim}, NEWEE \citep{Li2019}, EGAT \citep{Gong2019} that consume edge attributes.

\begin{table*}[ht]
    \caption{Comparison of most representative graph embedding methods with their abilities to learn the representation, with or without attributes, reasoning types and short characteristics. The most prominent and appropriate methods selected to compare to \texttt{AttrE2vec} in experiments are marked with bold text.}
    \label{tab:embedding-methods-comparison}
    \centering
    \scalebox{0.75}{
    \begin{tabular}{|c|l|cc|cc|cc|c|}
    \hline
    & \multirow{2}{*}{\textbf{Method}} & \multicolumn{2}{c|}{\textbf{Representation}} & \multicolumn{2}{c|}{\textbf{Attributed}} & \multicolumn{2}{c|}{\textbf{Reasoning}} & \multirow{2}{*}{\textbf{Family}} \\
   & & Nodes & Edges & Nodes & Edges & Transduct. & Induct. & \\
   \hline
   \parbox[t]{2mm}{\multirow{10}{*}{\rotatebox[origin=c]{90}{\textbf{Supervised}}}} & ECN \citep{Aggarwal2016a} (2016) &  & \checkmark &  &  & \checkmark &  & neigh. aggr. \\
  
   & GCN \citep{Kipf2019} (2017) & \checkmark &  & \checkmark &  & \checkmark & \checkmark & GCN/GNN\\
   & ECC \citep{Simonovsky} (2017) & \checkmark &  & \checkmark &  & \checkmark &  & GCN, DL\\
  
   & FSCNMF \citep{Bandyopadhyay2018} (2018) & \checkmark &  & \checkmark &  & \checkmark &  & GCN\\
   & GAT \citep{Velickovic2018} (2018) & \checkmark &  & \checkmark &  & \checkmark & \checkmark & AE, DL\\
   & Planetoid \citep{Bui2018} (2018) & \checkmark &  & \checkmark &  & \checkmark & \checkmark & GNN\\
  
   & EGNN \citep{Kim} (2019) & \checkmark & \checkmark & \checkmark & \checkmark & \checkmark & \checkmark & GNN\\
   & EdgeConv \citep{Wang2019a} (2019) &  & \checkmark &  &  & \checkmark &  & GNN\\
   & EGAT \citep{Gong2019} (2019) & \checkmark & \checkmark & \checkmark & \checkmark & \checkmark & \checkmark & GNN\\
  
   & Attribute2vec \citep{Wanyan2020} (2020) & \checkmark & \checkmark &  &  & \checkmark &  & GCN\\
   
   \hline
   
   \parbox[t]{2mm}{\multirow{13}{*}{\rotatebox[origin=c]{90}{\textbf{Unsupervised}}}} & \textbf{DeepWalk} \citep{Perozzia} (2014) & \checkmark &  &  &  & \checkmark &  & RW, skip-gram\\
  
   & TADW \citep{Yang2015} (2015) & \checkmark  & \checkmark &  &  & \checkmark &  & RW, MF \\
   & LINE \citep{Tang2015} (2015) & \checkmark &  &  &  & \checkmark &  & RW, skip-gram\\
   
   & \textbf{Node2vec} \citep{Grover2016} (2016) & \checkmark &  &  &  & \checkmark &  & RW, skip-gram\\
   & \textbf{SDNE} \citep{Wanga} (2016) & \checkmark &  & \checkmark &  & \checkmark & \checkmark & AE\\
   
   & \textbf{GraphSAGE} \citep{Hamiltona} (2017) & \checkmark &  & \checkmark &  & \checkmark & \checkmark & RW\\
   & EP-B \citep{Garcia-Duran2017} (2017) & \checkmark &  & \checkmark &  & \checkmark & \checkmark & AE\\
   & \textbf{Struc2vec} \citep{Ribeiro2017} (2017) & \checkmark &  &  &  & \checkmark &  & RW, skip-gram\\
   
   & DANE \citep{Gao2018a} (2018) & \checkmark &  & \checkmark &  & \checkmark & \checkmark & AE\\
   
   & \textbf{Line2vec} \citep{Bandyopadhyay} (2019) &  & \checkmark &  &  & \checkmark &  & RW, skip-gram\\
   & NEWEE \citep{Li2019} (2019) & \checkmark &  & \checkmark & \checkmark & \checkmark &  & RW, skip-gram\\
   
   \cline{2-9}
   
   & \textbf{AttrE2vec} (2020) &  & \checkmark & \checkmark & \checkmark & \checkmark & \checkmark & RW, AE, DL\\
   \hline
\end{tabular}
    }
\end{table*}

Both node-based embedding methods and graph neural network inspired methods do not generalize effectively to both transductive and inductive settings, especially when there are attributes associated with edges. This work is motivated by the idea of unsupervised learning on networks with attributed edges such that the embeddings are generalizable across tasks and are inductive.

%Motivation of our approach
To that end, we develop a novel \texttt{AttrE2vec}, an unsupervised learning model that adapts auto-encoder and self-attention network with the use of feature reconstruction and graph structural loss. To learn edge representation, \texttt{AttrE2vec} splits edge neighborhood into two parts, separately for each node endings of the edge, and then generates random edge walks in both neighborhoods. All walks are then aggregated over the node and edge attributes using one of the proposed strategies (Avg, Exp, GRU, ConcatGRU). These are accumulated with the original nodes and edge features and then fed to attention and dense layer to encode the edge. The embeddings are subsequently inferred via a two-step loss function --- for both feature reconstruction and graph structural loss. As a consequence, \texttt{AttrE2vec} can explicitly incorporate feature information from nodes and edges at many hops away to effectively produce the plausible edge embeddings for the inductive setting.

%Contributions
In summary, our main contributions are as follows:
\begin{itemize}
\item we propose a novel unsupervised \texttt{AttrE2vec} method, which learns a low-dimensional vector representation for edges that are attributed
\item we exploit the concept of a graph-topology-driven edge feature aggregation, from simple ones to learnable GRU based, that captures edge topological proximity and similarity of edge features
\item the proposed method is inductive and allows getting the representation for edges not present in the training phase 
\item we conduct various experiments and show that our \texttt{AttrE2vec} method has superior performance over all of the baseline methods on edge classification and clustering tasks.
\end{itemize}

\section{Related work and Research Gap}

Embedding information networks has received significant interest from the research community. We refer the readers to the survey articles for a comprehensive overview of network embedding \citep{Lia,Chami2020,Wu2019,Zhang2018a} and cite only some of the most prominent works that are relevant.

\textbf{Unsupervised network embedding methods} use only the network structure or original attributes of nodes and edges to construct embeddings. The most common method is DeepWalk \citep{Perozzia}, which in two-phases constructs node neighborhoods by performing fixed-length random walks and employs the skip-gram \citep{Grover2016} model to preserve the co-occurrences between nodes and their neighbors. This two-phase framework was later an inspiration for learning network embeddings by proposing different strategies for constructing node neighborhoods or modeling co-occurrences between nodes, e.g., node2vec \citep{Grover2016}, Struc2vec \citep{Ribeiro2017}, GraphSAGE \citep{Hamiltona}, line2vec \citep{Bandyopadhyay} or NEWEE \citep{Li2019}. Another group of unsupervised methods utilizes auto-encoder or graph neural networks to obtain embedding. SDNE \citep{Wanga} uses auto-encoder architecture to preserve first and second-order proximities by jointly optimizing the loss in neighborhood reconstruction. Another auto-encoder based representatives are EP-B \citep{Garcia-Duran2017} and DANE \citep{Gao2018a}. 

\textbf{Supervised network embedding methods} are constructed as an end-to-end methods for particular tasks like node classification or link prediction. These methods require network structure, attributes of nodes and edges (if method is capable of using) and some annotated target like node class. The representatives are ECN \citep{Aggarwal2016a}, ECC \citep{Simonovsky}, FSCNMF \citep{Bandyopadhyay2018}, GAT \citep{Velickovic2018}, planetoid \citep{Bui2018}, EGNN \citep{Kim}, GCN \citep{Kipf2019}, EdgeConv \citep{Wang2019a}, EGAT \citep{Gong2019}, Attribute2vec \citep{Wanyan2020}. 

\textbf{Edge representation learning} has been already tackled by several methods, i.e. 
ECN \citep{Aggarwal2016a}, EGNN \citep{Kim}, line2vec \citep{Bandyopadhyay}, EdgeConv \citep{Wang2019a}, EGAT \citep{Gong2019}. However, non of these methods was able to directly take into account attributes of edges as well as perform the learning in an unsupervised manner.

All the characteristics of the representative node and edge representation learning methods are grouped in Table \ref{tab:embedding-methods-comparison}.

\section{Method}

\subsection{Motivation}
In the following paragraphs, we explain our three-fold motivation to propose the \texttt{AttrE2vec}.

\paragraph{Edge embeddings} For a decade, network processing approaches gather more and more attention as graph data is produced in an increasing number of systems. Network embedding traditionally provided the notion of vectorizing nodes that was used in node classification or clustering. However, the edge representation learning did not gather enough attention and was accomplished through node embedding transformation \citep{node2vec}. Nevertheless, such an approach is problematic. For instance, inferring edge type from neighboring nodes' embeddings may not be the best choice for edge type classification in heterogeneous social networks. We claim that efficient edge clustering, edge attribute regression, or link prediction tasks require dedicated and specific edge representations. We expect that the representation learning approach devoted strictly to edges provides more powerful vector representations than traditional methods that require node embeddings trained upfront and transform nodes' embedding to represent edges.

\paragraph{Inductive embedding methods} A vast majority of contemporary network representation learning methods is transductive (see Table \ref{tab:embedding-methods-comparison}). It means that any change to the graph requires the whole retraining of the method to provide predictions for unseen cases—such property limits the applicability of methods due to high computational costs. Contrary, the inductive approach builds a predictive ability that can be applied to unseen cases and does not need retraining -- in general, inductive methods have a lower computation cost. Considering these advantages, we expect modern edge embedding methods to be inductive.

\paragraph{Encoding graph attributes in embeddings} Much of the real-world data exhibits rich attribute sets or meta-data that contain crucial information, e.g., about the similarity of nodes or edges. Traditionally, graph representation learning has been focused on exploiting the network structure, omitting the related content. Thus, we may expect to consume attributes as a regularizer over the structure. It would allow overcoming the limitation when the only edge discriminating ability is encoded in the edges' attributes, not in the graph's structure. Relying only on the network would produce inconclusive embeddings.

\subsection{Attributed graph edge embedding}
We denote an attributed graph as $G = (V, E)$, where $V$ is a set of nodes and $E = \{(u, v) \in V \times V\}$ a set of edges. Every node $u$ and every edge $e = (u, v)$ has associated features: $m_u \in \mathbb{R}^{d_V}$ and $f_{uv} \in \mathbb{R}^{d_E}$, where $\mathcal{M} \in \mathbb{R}^{|V| \times d_V}$ and $\mathcal{F} \in \mathbb{R}^{|E| \times d_E}$ are node and edge feature matrices, respectively. By $d_V$ we denote dimensionality of node feature space and $d_E$ dimensionality of edge feature space. The edge embedding task is defined as learning a function $g: E \to \mathbb{R}^{d}$, which takes an edge and outputs its low-dimensional vector representation. Note that the embedding dimension $d$ should be much less than the original edge feature dimensionality $d_E$, i.e.: $d << d_E$. More specifically, we aim at using the topological structure of the graph and node and edge attributes: $f: (E, \mathcal{F}, \mathcal{M}) \to \mathbb{R}^d$.

\begin{figure*}[ht]
  \centering
  \includegraphics[width=\textwidth]{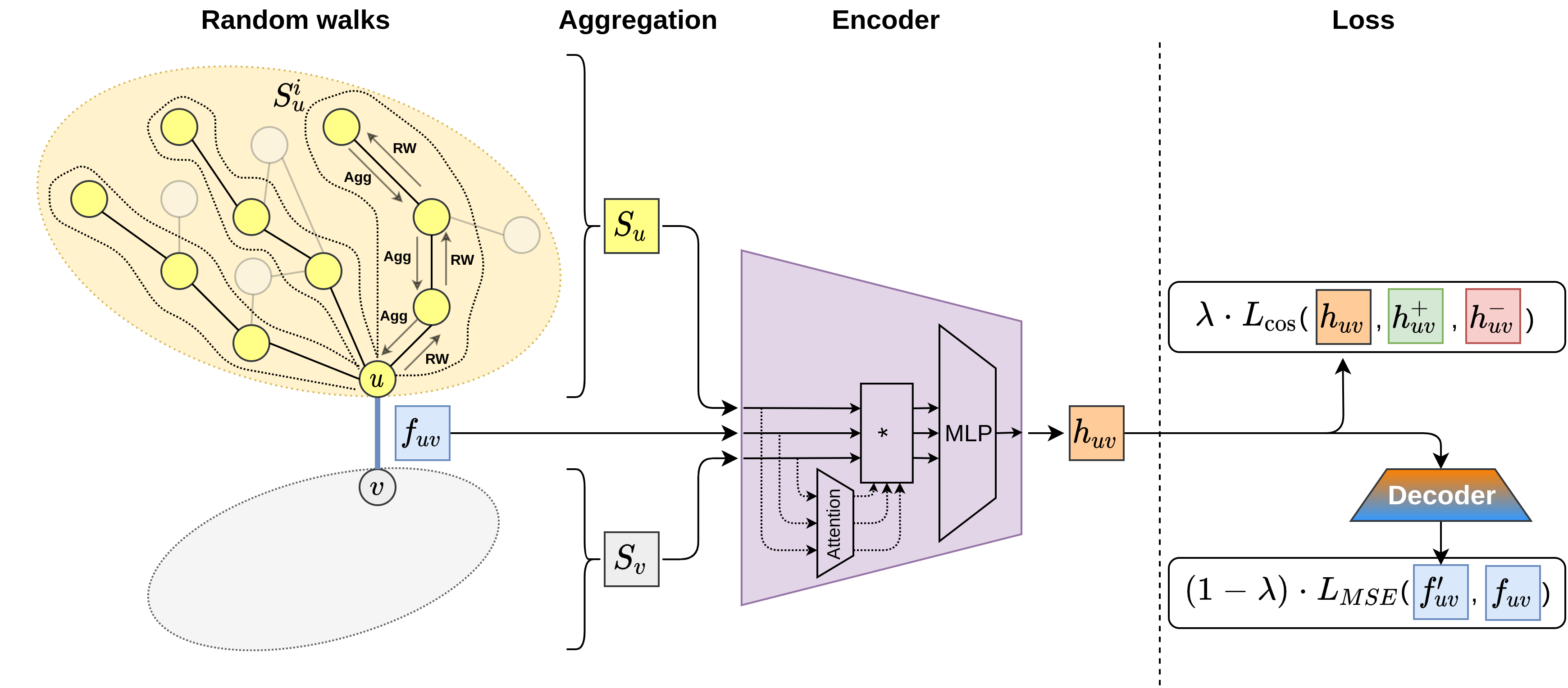}
  \caption{Overview of the \texttt{AttrE2vec} model. The model first computes edge random walks on two neighborhoods of a given edge $(u, v)$. Each neighbourhood walks are aggregated into $S_u, S_v$. Both are combined with the edge features $f_{uv}$ using an Encoder module, which results into the edge embedding vector $h_{uv}$. The loss function consists of two parts: structural loss ($L_{\text{cos}}$) and feature reconstruction loss ($L_{\text{MSE}}$). }
  \label{fig:attre2vec-model}
\end{figure*}

\subsection{AttrE2vec}
In contrast to traditional node embedding methods, we shift the focus from nodes to edges and consider a graph from an edge perspective. Given any edge $e = (u, v)$, we can observe three natural sources of knowledge: the edge attributes itself and the two neighborhoods - $N_u$ and $N_v$, located behind nodes $u$ and $v$, respectively. In AttrE2vec, we exploit all three sources jointly. 

First, we obtain aggregations (summaries) $S_u, S_v$ of the both neighborhoods $N_u, N_v$. We want to capture the topological structure of the neighborhood, so we perform $k$ \textbf{edge random walks} of length $L$, which start from node $u$ (or $v$, respectively) and use a uniformly distributed neighbor sampling approach (DeepWalk-like) to obtain the next edge. Each $i$th walk $w_u^i$ started from node $u$ is hence a sequences of edges.

$$\mathbf{RW}(G, k, L, u) \to \{w_{u}^1, w_{u}^2, \ldots, w_{u}^k\}$$
$$w_{u}^i \equiv (u, u_2), (u_3, u_4), \ldots, (u_{L-1}, u_L)$$

Next, we take the attributes of the edges (and nodes, if applicable) in each random walk and aggregate them into a single vector using the \textbf{walk aggregation model} $\mathbf{Agg_w}$.

$$S_u^i = \mathbf{Agg_w}(w_u^i, \mathcal{F}, \mathcal{M})$$

Later, aggregated walks are combined using the \textbf{neighborhood aggregation model} $\mathbf{Agg_n}$, which summarizes the neighborhood $S_u$ (and $S_v$, respectively). The proposed implementations of these aggregation are given in Section \ref{sec:agg-models}.

$$S_u = \mathbf{Agg_n}(\{S_u^1, S_u^2, \ldots, S_u^k\})$$

\begin{figure}[H]
    \centering
    \includegraphics[width=0.9\columnwidth]{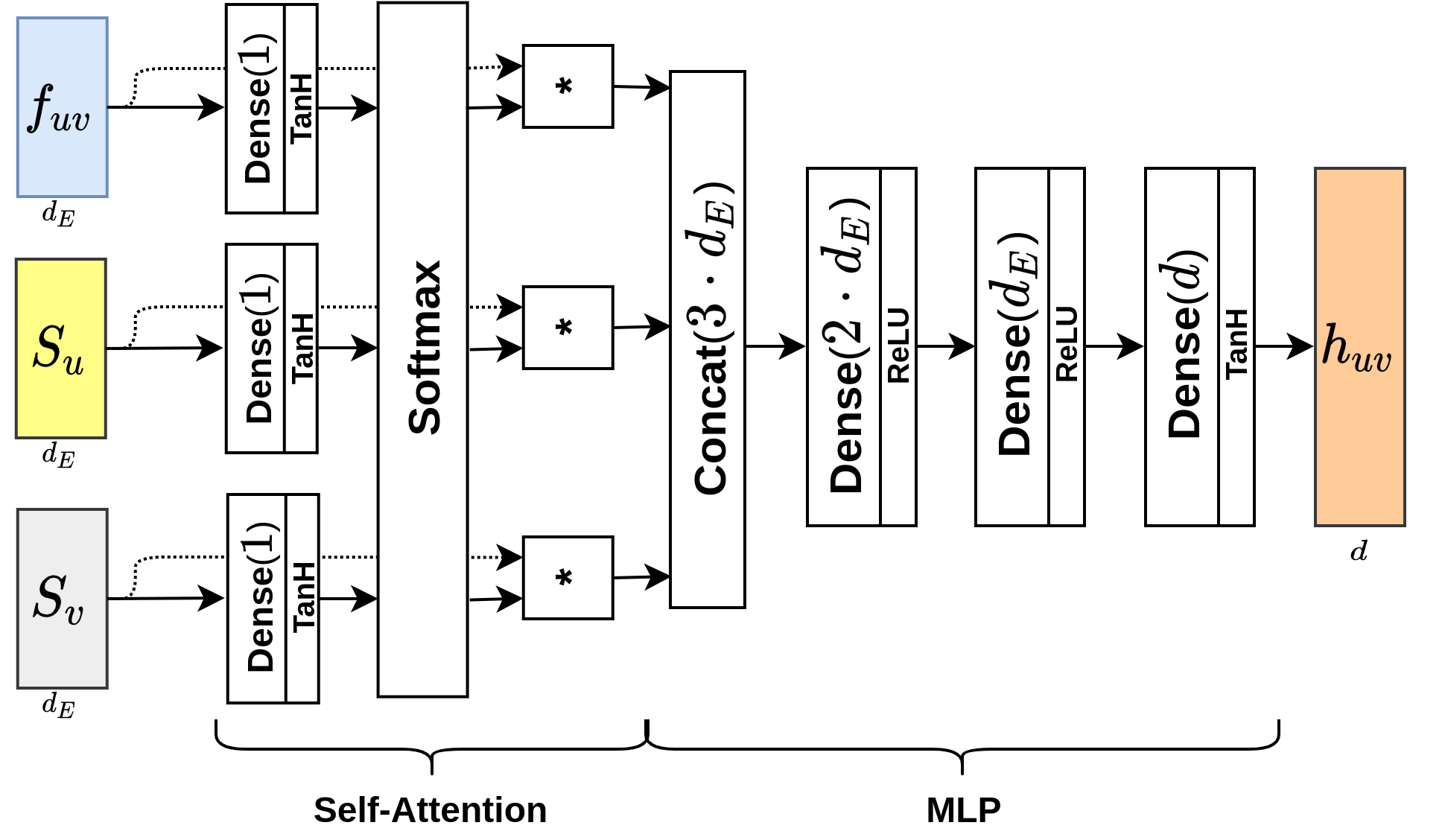}
    \caption{Encoder module architecture}
    \label{fig:encoder-architecture}
\end{figure}

Finally, we obtain the low dimensional edge embedding $h_{uv}$ using an \textbf{encoder} $\mathbf{Enc}$ module. It combines the edge attributes $f_{uv}$ with the summarized neighborhood information $S_u$, $S_v$. We employ a simple Multilayer Perceptron (MLP) with 3 inputs (each of size equal to the edge features dimensionality) and an attention mechanism over these inputs, to check how much of the information of each input is used to create the embedding vector (see Figure \ref{fig:encoder-architecture}):

$$h_{uv} = \mathbf{Enc}(f_{uv}, S_u, S_v)$$

The overall illustration of the method is contained in Figure \ref{fig:attre2vec-model} and the inference algorithm is shown in Algorithm \ref{alg:attre2vec-inference}. 

\begin{algorithm}[ht]
    \SetAlgoLined
    \DontPrintSemicolon
    
    \KwData{graph $G$, edge list $xe$, edge features $\mathcal{F}$, node features $\mathcal{M}$}
    \kwParams{number of random walks per node $k$, random walk length $L$}
    \KwResult{edge embedding vectors $h_{uv}$}

    \Begin{
        \ForEach{($u$, $v$) in $xe$}{
            \ForEach{$i$ in (1\ldots k)}{
                $w_u^i = \mathbf{RW}(G, L, u)$\;
                $S_u^i = \mathbf{Agg_w}(w_u^i,\mathcal{F}, \mathcal{M})$\;
                \;
                $w_v^i = \mathbf{RW}(G, L, v)$\;
                $S_v^i = \mathbf{Agg_w}(w_v^i,\mathcal{F}, \mathcal{M})$\;
            }
            $S_u = \mathbf{Agg_n}(\{S_u^1,\ldots,S_u^k\})$\;
            $S_v = \mathbf{Agg_n}(\{S_v^1,\ldots,S_v^k\})$\;
            \;
            $h_{uv} = \mathbf{Enc}(f_{uv}, S_u, S_v)$\;
        }
    }
    \caption{AttrE2vec inference algorithm}
    \label{alg:attre2vec-inference}
\end{algorithm}

\subsection{Aggregation models}\label{sec:agg-models}
For the purpose of the neighborhood aggregation model $\mathbf{Agg_n}$, we use an average over vectors $S_u^i$, as there is no particular ordering of these vectors (each one was generated by an equally important random walk). In the case of walk aggregation, we propose the following:
\begin{itemize}
    \item \textbf{average} -- that computes a simple average of the edge attribute vectors in the random walk;
            $$S_u^i = \frac{1}{L} \sum_{n = 1}^L f_{u_{n}u_{n+1}} $$
    
    \item \textbf{exponential} -- that computes a weighted average, where the weights are $exp$onents of the "minus" position in the random walk so that further away edges are less important than the near ones;
            $$S_u^i = \frac{1}{L} \sum_{n = 1}^L e^{-n} f_{u_{n}u_{n+1}}$$
    
    \item \textbf{GRU} -- that uses a Gated Recurrent Unit \citep{Chung2014EmpiricalModeling} architecture, where hidden and input dimension is equal to the edge attribute dimension; the aggregated representation is the output of the last hidden vector; the aggregation process starts here at the end of the random walk and proceeds to the beginning;
            $$S_u^i = \text{GRU}(\{f_{u_{n}u_{n+1}}, f_{u_{n-1}u_{n}}, \ldots, f_{u_{1}u_{2}}\})$$

    \item \textbf{ConcatGRU} -- that is similar to the GRU-based aggregator, but here we also use the node feature information by concatenating the node attributes with the edge attributes; hence the GRU input size is equal to the sum of the edge and node dimensions; in case there are not any node features available, one could use network-specific features, like degree, betweenness or more advanced techniques like Node2vec; the hidden dimension size and the aggregation direction is unchanged;
            $$S_u^i = \text{ConcatGRU}(\{f_{u_{n}u_{n+1}} \oplus m_{u_{n}}, \ldots, f_{u_{1}u_{2}} \oplus m_{u_1}\})$$

\end{itemize}

\subsection{Learning AttrE2vec's parameters}
\begin{figure}[H]
    \centering
    \includegraphics[width=0.5\columnwidth]{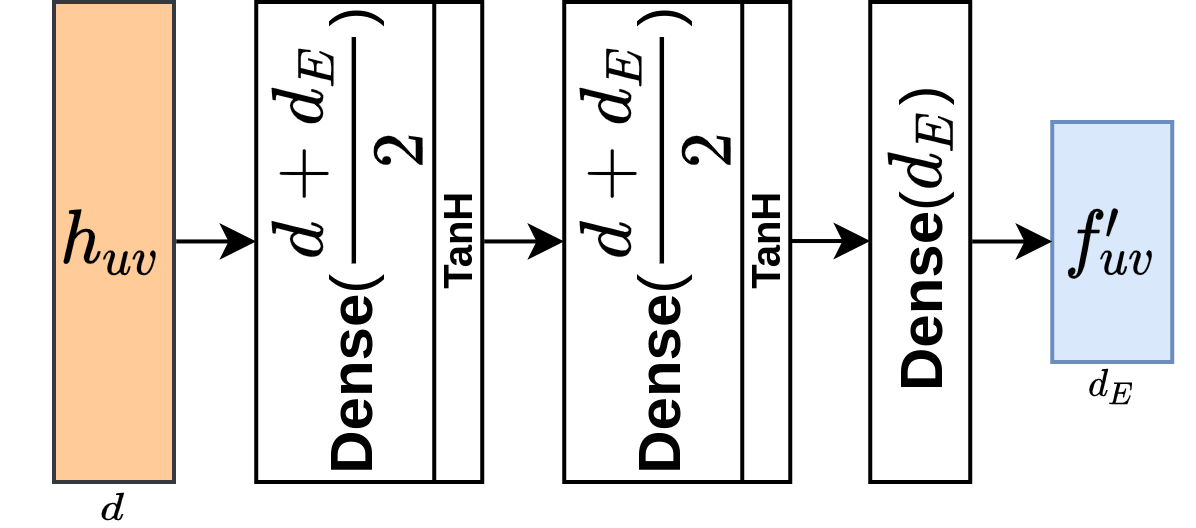}
    \caption{Decoder module architecture}
    \label{fig:decoder-architecture}
\end{figure}

\texttt{AttrE2vec} is designed to make the most use of edge attributes and information about the structure of the network. Therefore we propose a loss function, which consists of two main parts:
\begin{itemize}
    \item structural loss $L_{\text{cos}}$ -- computes a \textbf{cosine embedding loss}; such function tries to minimize the cosine distance between a given embedding $h$ and embeddings of edges sampled from the random walks $h^+$ (positive), and simultaneously to maximize a cosine distance between an embedding $h$ and embeddings of edges sampled from a set of all edges in the graph $h^-$ (negative), except for these in the random walks:

    \begin{equation*}
    \displayindent0pt
    \displaywidth\columnwidth
        L_{\text{cos}} = \frac{1}{|B|} \sum_{h_{\text{uv}} \in B} \left( \sum_{h_{\text{uv}}^+} (1 - \cos(h_{\text{uv}}, h_{\text{uv}}^+)) + \sum_{h_{\text{uv}}^-} \cos(h_{\text{uv}}, h_{\text{uv}}^-) \right)
    \end{equation*}
    where $B$ denotes a minibatch of edges and $|B|$ the minibatch size,
    
    \item feature reconstruction loss $L_{\text{MSE}}$ -- computes a \textbf{mean squared error} of the actual edge features and the outputs of a \textbf{decoder} (implemented as a 3-layer MLP -- see Figure \ref{fig:decoder-architecture}), that reconstruct the edge features based on the edge embeddings;
    
    \begin{equation*}
        L_{\text{MSE}} = \frac{1}{|B|} \sum_{(h_{\text{uv}}, f_{\text{uv}}) \in B} \left( \text{DEC}(h_{\text{uv}}) - f_{\text{uv}} \right)^2
    \end{equation*}
    where $B$ denotes a minibatch of edges and $|B|$ the minibatch size.
\end{itemize}

We combine the values of the above loss functions using a mixing parameter $\lambda \in [0,1]$. The  higher the value of this parameter is, the more structural information is preserved and less focus is one the feature reconstruction. The total loss of AttrE2vec is given as follows:

\begin{equation*}
    L = \lambda * L_{\text{cos}} + (1 - \lambda) * L_{\text{MSE}}
\end{equation*}

\section{Experiments}

To evaluate the proposed model's performance, we perform three tasks: edge classification, edge clustering, and embedding visualization on three real-world datasets.  
We first train our model on a small subset of edges (inductive setting). Then we use the model to infer embeddings for edges from the test set. Finally, we evaluate them in all downstream tasks: by predicting the class of edges in citation graphs (\textbf{edge classification}), by applying the K-means++ algorithm (\textbf{edge clustering}; as defined in \citep{Bandyopadhyay}) and by the dimensionality reduction method T-SNE (\textbf{embedding visualization}). We compare our model to several baselines and contemporary methods in all experiments, see Table \ref{tab:embedding-methods-comparison}. Eventually, we check the influence of AttrE2vec's hyperparameters and perform an ablation study on artificially generated datasets. We implement our model in the popular deep learning framework PyTorch. All experiments were performed on an NVIDIA GTX1080Ti. Upon acceptance in the journal, we will make our code available at \texttt{\url{https://github.com/attre2vec/attre2vec}} and include our DVC \citep{dvc} pipeline so that all experiments can be easily reproduced.

\subsection{Datasets}\label{sec:exp-datasets}
\begin{table*}[ht]
    \renewcommand{\arraystretch}{1.3}
    \caption{Datasets used in the experiments.}
    \label{tab:datasets}
    \centering
    
    \scalebox{0.8}{
    \begin{tabular}{|c||cc|cc||ccc||cc|}
        \hline
        \multirow{3}{*}{\textbf{Name}} & \multicolumn{4}{c||}{\textbf{Features}} & \multicolumn{3}{c||}{\multirow{2}{*}{\textbf{Number of}}} & \multicolumn{2}{c|}{\multirow{2}{*}{\textbf{Training instances}}}\\ \cline{2-5}
        & \multicolumn{2}{c|}{\textbf{initial}} & \multicolumn{2}{c||}{\textbf{pre-processed}} & \multicolumn{3}{c||}{} & \multicolumn{2}{c|}{} \\ \cline{2-10}
        & \textbf{node} & \textbf{edge} & \textbf{node} & \textbf{edge} & \textbf{nodes} & \textbf{edges} & \textbf{classes} & \textbf{inductive} & \textbf{transductive}\\ \hline
        Cora & 1 433 & 0 & 32 & 260 & 2 485 & 5 069 & 7+1 & 160 & 5 069\\\hline
        Citeseer & 3 703 & 0 & 32 & 260 & 2 110 & 3 668 & 6+1 & 140 & 3 668\\\hline
        Pubmed & 500 & 0 & 32 & 260 & 19 717 & 44 324 & 3+1 & 80 & 44 324\\\hline
    \end{tabular}
    }
\end{table*}

In order to compare gathered evaluation evidence we focused on well known datasets, that appear in the literature, namely: Cora \citep{cora_dataset}, Citeseer \citep{cora_dataset} and Pubmed \citep{pubmed_dataset}. These are citation networks of scientific papers in several research areas, where nodes are the papers and edges denote citations between papers. We summarize basic statistics about the datasets before and after pre-processing steps in Table \ref{tab:datasets}. Raw datasets contain node features only in the form of high dimensional sparse bags of words. For Cora and Citeseer, these are binary vectors, showing which of the most popular words were used in a given paper, and for Pubmed, the features are in the form of TF-IDF vectors. To adjust the datasets to our problem setting, we apply the following pre-processing steps to obtain edge level features, which are used to train and evaluate our \texttt{AttrE2vec} model:
\begin{itemize}
    \item we create dense vector representations of the nodes' features by applying Doc2vec \citep{Le} in the PV-DBOW variant with a target dimension size of 128;
    \item for each edge $(u, v)$ and its symmetrical version $(v, u)$ (necessary to perform uniform, undirected random walks) we extract the following features:
    \begin{itemize}
        \item 1 feature -- cosine similarity of raw node features for nodes $u$ and $v$ (binary BoW; for Pubmed transformed from TF-IDF to binary BoW),
        \item 2 features -- the ratios of the number of used words (number of ones in the BoW) to all possible words in the document (length of BoW vector) in each paper $u$ and $v$,
        \item 256 features -- concatenation of Doc2vec features for nodes $u$ and $v$,
        \item 1 feature -- a binary indicator, which denotes whether this is an original edge (1) or its symmetrical counterpart (0),
    \end{itemize}
    \item we apply standardization (StandardScaler in Scikit-Learn \citep{scikit-learn}) of the edge feature matrix.
\end{itemize}

Moreover, we extracted new node features as 32-dimensional Node2vec embeddings to provide the evaluation possibility for one of our model versions (AttrE2vec with ConcatGRU aggregator), which generalizes upon both edge and nodes attributes.

Raw datasets provide each node labeled by the research area the paper comes from. To apply this knowledge in the edge classification problem setting, we applied the following rule: if an edge has two nodes from the same class (research area), the edge receives this class; if two nodes have different classes, the edge between these nodes is assigned with a cross-domain citation class.

To ensure a fair comparison method, we follow the dataset preparation scheme from EP-B \citep{Garcia-Duran2017}, i.e., for each dataset (Cora, Citeseer, Pubmed) we sample 10 train/validation/test sets, where the train set consists of 20 edges per class and the validation and test sets to contain 1 000 randomly chosen edges each. While reporting the resulting metrics, we show the mean values over these ten sampled sets (together with the standard deviation).

\subsection{Baselines}
We compare our method against several baseline methods. In the most simple case, we use the edge features obtained during the pre-processing phase for all datasets (further referred to as \textit{Doc2vec}). 

Many standard approaches employ simple node embedding transformations to obtain edge embeddings. The authors of Node2vec \citep{node2vec} proposed binary operators like averaging, Hadamard product, or L1 and L2 norms of vector differences. Here, we will use following methods to obtain node embeddings: DeepWalk \citep{Perozzia}, Node2vec \citep{node2vec}, SDNE \citep{sdne} and Struc2vec \citep{Ribeiro2017}. In preliminary experiments, we evaluated these methods and checked that the Average operator and an embedding size of 64 gives the best results. We will use these models in 2 setups: (a) \textbf{Avg($\mathcal{M}$,$\mathcal{M}$)} -- using only the averaged node features, (b) \textbf{Avg($\mathcal{M}$,$\mathcal{M}$)$\oplus\mathcal{F}$}  -- like previously but concatenated with the edge features from the dataset (in total 324-dim vectors). 

We also checked a scheme to compute a 64-dim PCA reduction of the concatenated features to have comparable vector sizes with the 64-dimensional embedding of our model, but these turned out to perform poorly. Note that SDNE has the capability of inductive reasoning, but due to the non-availability of such implementation, we decided to evaluate this method in the transductive scheme (which works in favor of the method). 

\begin{figure}[H]
    \centering
    \includegraphics[width=0.6\textwidth]{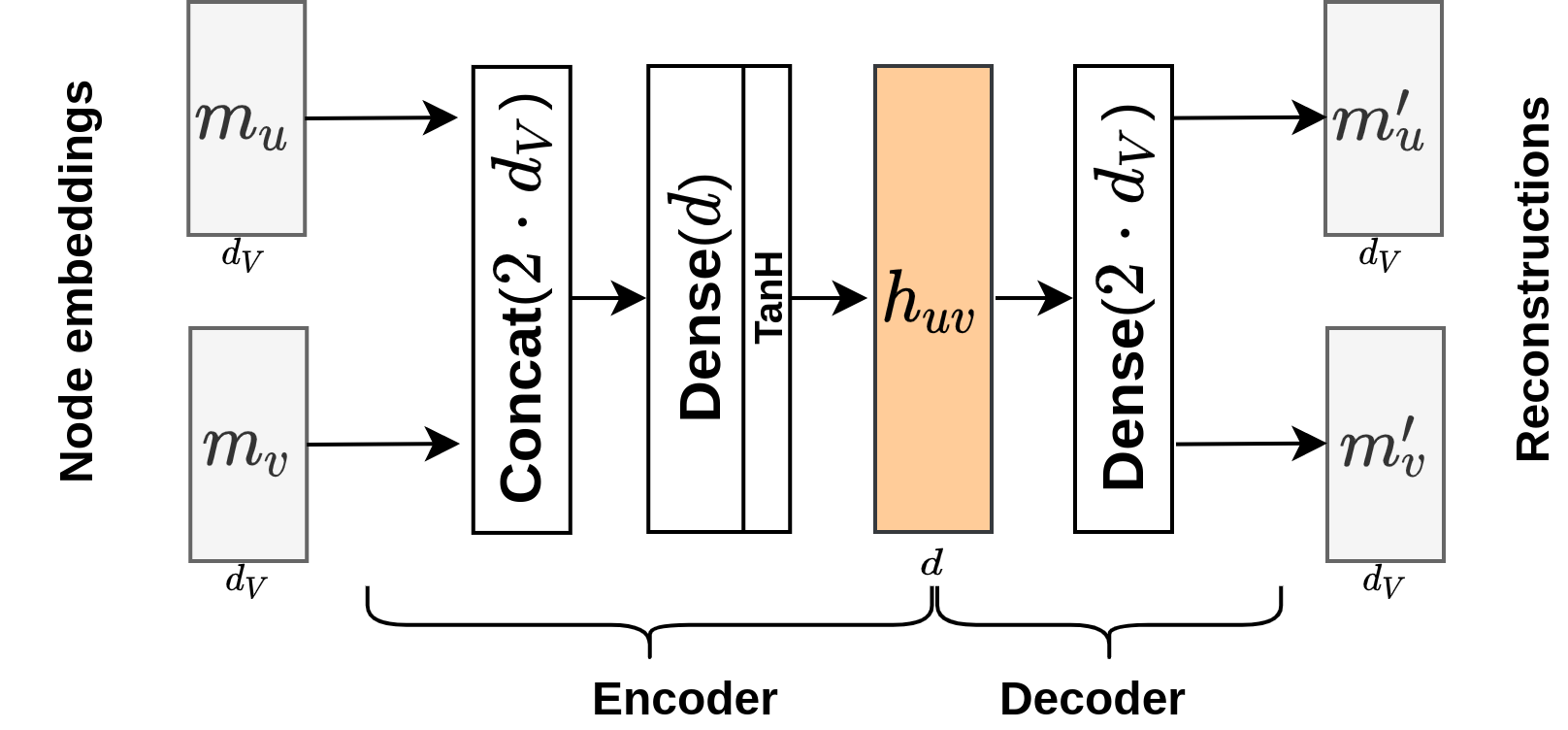}
    \caption{Architecture of the \textbf{MLP($\mathcal{M}$,$\mathcal{M}$)}.}
    \label{fig:mlp2-model}
\end{figure}

\begin{figure}[H]
    \centering
    \includegraphics[width=0.7\textwidth]{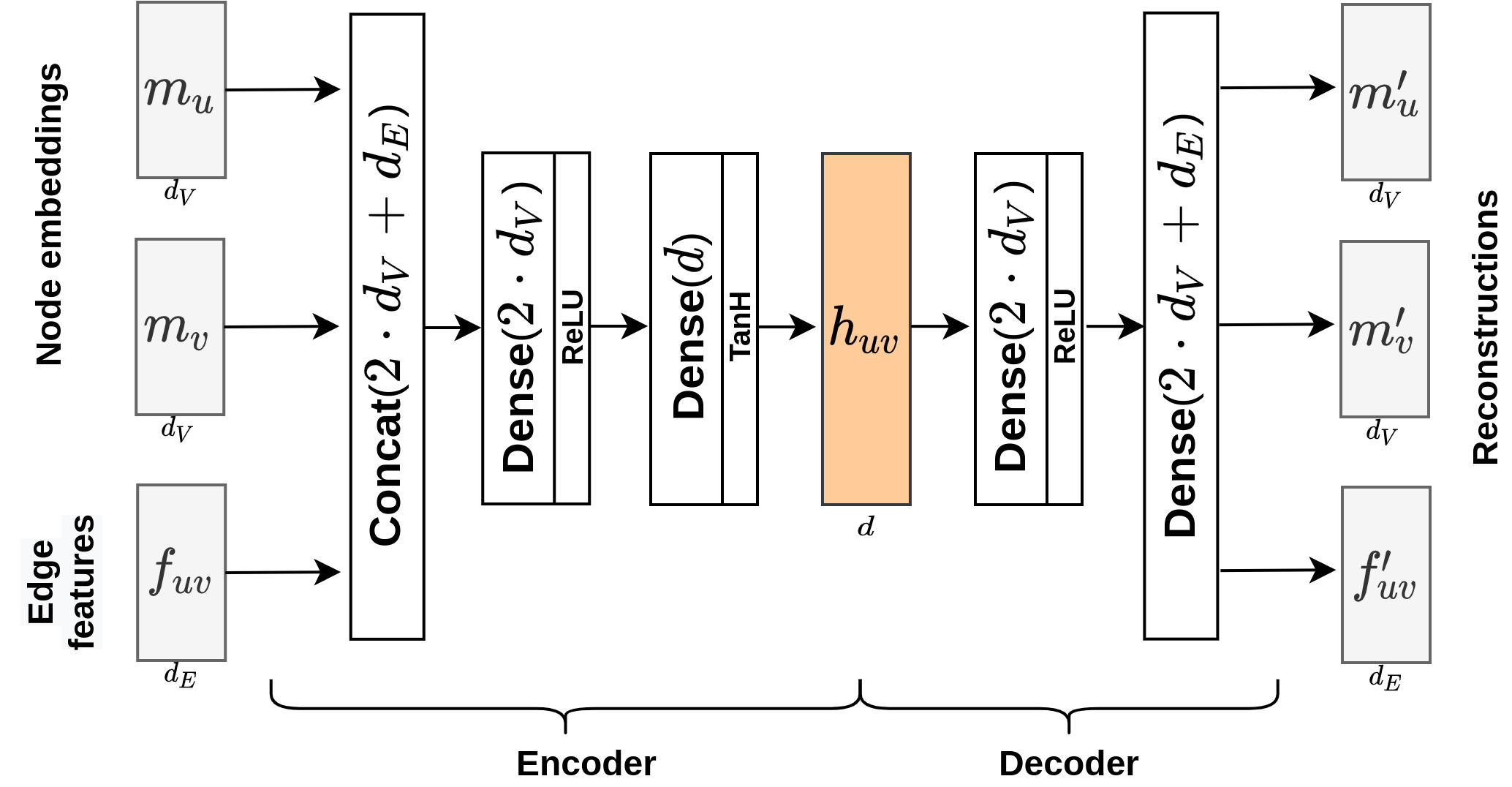}
    \caption{Architecture of the \textbf{MLP($\mathcal{M}$,$\mathcal{M}$,$\mathcal{F}$)}.}
    \label{fig:mlp3-model}
\end{figure}

We also extend our body of baselines by more sophisticated approaches -- two dense autoencoder architectures. In the first setting \textbf{MLP($\mathcal{M}$,$\mathcal{M}$)}, we train a model (see Figure \ref{fig:mlp2-model}), which reconstructs concatenated embeddings of connected nodes. In the second baseline \textbf{MLP($\mathcal{M}$,$\mathcal{M}$,$\mathcal{F}$)}, the autoencoder (see Figure \ref{fig:mlp3-model}) is extended by edge attributes. In both settings, we employ the mean squared error as the model loss function. The output of the encoders (embeddings) is used in the downstream tasks. The input node embeddings are obtained using the methods mentioned above, i.e., DeepWalk, Node2vec, SDNE, and Struc2vec.

The last baseline is Line2vec \citep{Bandyopadhyay}, which is directly dedicated for edges - we use an embedding size of 64.

\begin{table*}[ht]
    \caption{AUC values for edge classification. $\mathcal{F}$ denotes the edge attributes (also referred to as "Doc2vec"), $\mathcal{M}$ -- node attributes  (e.g., embeddings computed using "Node2vec"), $\oplus$ -- concatenation operator, \textbf{Avg($\mathcal{M}$,$\mathcal{M}$)} -- average operator on node embeddings, \textbf{MLP($\cdot$)} -- encoder output of MLP autoencoder trained on given attributes. \textbf{AUC} in bold shows the highest value and \textit{AUC} in italic --- the second highest value.}
    \label{tab:edge-clf-metrics}
    \centering
    \scalebox{0.85}{
        \begin{tabular}{|l|l|l|c|ccc|}
\hline
& \multicolumn{2}{c|}{\multirow{2}{*}{\textbf{Method group/name}}} & \textbf{Vector} & \multicolumn{3}{c|}{\textbf{AUC}}  \\
& \multicolumn{2}{c|}{$\;$}  & \textbf{size}  & \textbf{Citeseer} & \textbf{Cora} & \textbf{Pubmed} \\

\hline

\parbox[t]{2mm}{\multirow{18}{*}{\rotatebox[origin=c]{90}{\textbf{Transductive}}}} & \multicolumn{2}{c|}{\textbf{Edge features only; $\mathcal{F}$ } (Doc2vec)} & 260 &  86.13 $\pm$ 0.95 &  88.67 $\pm$ 0.51 &  79.15 $\pm$ 1.41 \\
\cline{2-7}
& \multicolumn{2}{l|}{\textbf{Line2vec}} & 64 &  86.19 $\pm$ 0.28 &  91.75 $\pm$ 1.07 &  84.88 $\pm$ 1.19 \\
\cline{2-7}

& \multirow{4}{*}{\textbf{Avg($\mathcal{M}$,$\mathcal{M}$)}} & DeepWalk & 64 & 58.40 $\pm$ 1.08 &  59.98 $\pm$ 1.32 &  51.04 $\pm$ 1.23 \\
& & Node2vec   & 64 &  58.26 $\pm$ 0.89 &  59.59 $\pm$ 1.11 &  51.03 $\pm$ 1.01 \\
& & SDNE       & 64 &  54.28 $\pm$ 1.57 &  55.91 $\pm$ 1.11 &  50.00 $\pm$ 0.00 \\
& & Struc2vec  & 64 &  61.29 $\pm$ 0.86 &  61.30 $\pm$ 1.58 &  54.67 $\pm$ 1.46 \\
\cline{2-7}

& \multirow{4}{*}{\textbf{MLP($\mathcal{M}$,$\mathcal{M}$)}} & DeepWalk & 64 & 55.88 $\pm$ 1.68 &  57.87 $\pm$ 1.53 &  51.23 $\pm$ 0.77 \\
& & Node2vec & 64 & 55.35 $\pm$ 2.26 &  57.44 $\pm$ 0.87 &  51.48 $\pm$ 1.55 \\
& & SDNE & 64 & 55.56 $\pm$ 0.93 &  56.02 $\pm$ 1.22 &  50.00 $\pm$ 0.00 \\
& & Struc2vec & 64 & 59.93 $\pm$ 1.43 &  59.76 $\pm$ 1.80 &  53.27 $\pm$ 1.32 \\

\cline{2-7}

& \multirow{4}{*}{\textbf{Avg($\mathcal{M}$,$\mathcal{M}$)}$\oplus\mathcal{F}$}& DeepWalk & 324 &  86.13 $\pm$ 0.95 &  88.67 $\pm$ 0.51 &  79.15 $\pm$ 1.41 \\
& & Node2vec  & 324 &  86.13 $\pm$ 0.95 &  88.67 $\pm$ 0.51 &  79.15 $\pm$ 1.41 \\
& & SDNE      & 324 &  86.14 $\pm$ 1.03 &  88.70 $\pm$ 0.51 &  79.15 $\pm$ 1.41 \\
& & Struc2vec & 324 &  86.21 $\pm$ 0.97 &  88.73 $\pm$ 0.48 &  79.24 $\pm$ 1.36 \\

\cline{2-7}

& \multirow{4}{*}{\textbf{MLP($\mathcal{M}$,$\mathcal{M}$,$\mathcal{F}$)}} & DeepWalk & 64 & 84.58 $\pm$ 1.11 &  86.47 $\pm$ 0.87 &  78.60 $\pm$ 1.84 \\
& & Node2vec & 64 & 84.65 $\pm$ 1.05 &  86.71 $\pm$ 0.68 &  78.84 $\pm$ 1.71 \\
& & SDNE & 64 &  84.32 $\pm$ 1.13 &  85.99 $\pm$ 0.77 &  78.34 $\pm$ 1.07 \\
& & Struc2vec & 64 & 83.95 $\pm$ 1.16 &  85.54 $\pm$ 0.96 &  77.19 $\pm$ 1.42 \\

\hline\hline

\parbox[t]{2mm}{\multirow{8}{*}{\rotatebox[origin=c]{90}{\textbf{Inductive}}}} & \textbf{Avg($\mathcal{M}$,$\mathcal{M}$)} & GraphSage & 64 & 54.84 $\pm$ 1.90 & 55.16 $\pm$ 1.36 & 51.14 $\pm$ 1.64 \\
\cline{2-7}
& \textbf{MLP($\mathcal{M}$,$\mathcal{M}$)} & GraphSage & 64 &  55.19 $\pm$ 1.04 &  55.47 $\pm$ 1.66 &  50.36 $\pm$ 1.54 \\
\cline{2-7}

& \textbf{Avg($\mathcal{M}$,$\mathcal{M}$)$\oplus\mathcal{F}$} & GraphSage & 324 & 86.14 $\pm$ 0.95 & 88.68 $\pm$ 0.51 & 79.16 $\pm$ 1.41 \\
\cline{2-7}

& \textbf{MLP($\mathcal{M}$,$\mathcal{M}$,$\mathcal{F}$)} & GraphSage & 64 &  84.63 $\pm$ 1.11 &  86.14 $\pm$ 0.45 &   78.00 $\pm$ 1.85 \\
\cline{2-7}

 & \multirow{4}{*}{\textbf{AttrE2vec} (our)} & Avg & 64 & $\mathbf{88.97 \pm 0.82}$ &  $\mathbf{93.43 \pm 0.56}$ &  $\mathbf{87.68 \pm 1.25}$ \\
& & Exp       & 64 &  88.91 $\pm$ 1.10 &  92.80 $\pm$ 0.38 &  86.18 $\pm$ 1.41 \\
& & GRU       & 64 & $\mathit{88.92 \pm 1.13}$ & $\mathit{93.06 \pm 0.63}$ &  $\mathit{86.39 \pm 1.21}$ \\
& & ConcatGRU & 64 &  88.56 $\pm$ 1.34 &  92.93 $\pm$ 0.61 &  86.34 $\pm$ 1.18 \\

\hline

\end{tabular}
    }
\end{table*}

\subsection{Edge classification}
To evaluate our model in an inductive setting, we need to make sure that test edges are unseen during the model training procedure -- we remove them from the graph. Note that all baselines (except for GraphSage, see \ref{tab:embedding-methods-comparison}) require all edges during the training phase (i.e., these are transductive methods).

After each training epoch of \texttt{AttrE2vec}, we evaluate the embeddings using L2-regularized Logistic Regression (LR) classifier and compute AUC. The regression model is trained on edge embeddings from the train set and evaluated on edge embeddings from the validation set. We take the model with the highest AUC value on the validation set.  Moreover, an early stopping strategy is implemented-- if the validation AUC metric does not improve for more than 15 epochs, the learning is terminated. Our approach to model selection is aligned with the schema proposed in \citep{QuocNguyenAModel} because this approach is more natural than relying on the loss function.  This is repeated for all 10 data splits (see: Section \ref{sec:exp-datasets} for details). We report a mean and std AUC measures for 10 test sets (see Table \ref{tab:edge-clf-metrics})

We choose AdamW \citep{Loshchilov} with a learning rate of $0.001$ to optimize our model's parameters. We also set the size of positive samples to $|h^+| = 5$ and negative samples to $|h^-|= 10$ in the cosine embedding loss. The mixing coefficient is set to $\lambda = 0.5$, equally including the influence of features and topological graph structure. We choose an embedding size of 64 as a reasonable value while dealing with edge features of size 260.

% Discussion
In Table \ref{tab:edge-clf-metrics}, we summarize the AUC values for baseline methods and for our model. Even though vectors' original dimensionality is relatively high (260), good results are already yielded using only the edge features (Doc2vec). However, adding structural information about the graph could further improve the results.

Using representations from node embedding methods, which are transformed to edge embeddings using the average operator \textbf{Avg($\mathcal{M}$,$\mathcal{M}$)}, achieve poor results of about 50-60\% AUC. However, if these are combined with the edge features from the datasets \textbf{Avg($\mathcal{M}$,$\mathcal{M}$)$\oplus\mathcal{F}$}, the AUC values increase significantly to about 86\%, 88\% and 79\% for Citeseer, Cora, and Pubmed, respectively. Unfortunately, this results in an even higher vector dimensionality (324).

The MLP-based approach results lead to similar conclusions. Using only node embeddings \textbf{MLP($\mathcal{M}$,$\mathcal{M}$)} we achieve quite poor results of about 50\% (on Pubmed) up to 60\% (on Cora). With \textbf{MLP($\mathcal{M}$,$\mathcal{M}$,$\mathcal{F}$)} approach we observe that edge features improve the classification results. The AUC values are still slightly worse than concatenation operator (\textbf{Avg($\mathcal{M}$,$\mathcal{M}$)$\oplus\mathcal{F}$}), but we can reduce the edge embedding size to 64.

The Line2vec \citep{Bandyopadhyay} algorithm achieves very good results, without considering edge features information -- we get about 86\%, 92\% and 85\% AUC for Citeseer, Cora, and Pubmed, respectively. These values are higher than for any other baseline approach. 

Our model performs the best among all evaluated methods. For Citeseer, we gain about 3 percent points compared to the best baselines: Line2vec, Struc2vec (\textbf{Avg($\mathcal{M}$,$\mathcal{M}$)$\oplus\mathcal{F}$}) or GraphSage (\textbf{Avg($\mathcal{M}$,$\mathcal{M}$)$\oplus\mathcal{F}$}). Note that the algorithm is trained only on 140 edges in the inductive setting, whereas all transductive baselines require the whole graph for training. The gains on Cora are 2 pp, and on Pubmed we achieve up to 4pp (and up to 8pp compared only to GraphSage (\textbf{Avg($\mathcal{M}$,$\mathcal{M}$)$\oplus\mathcal{F}$)}). Our model with the Average (Avg) aggregator works the best, whereas the Gated Recurrent Unit (GRU) aggregator achieves the second-best results.

\begin{figure*}
  \centering
  \includegraphics[width=0.3\textwidth]{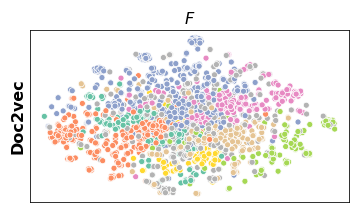}%
  \includegraphics[width=0.3\textwidth]{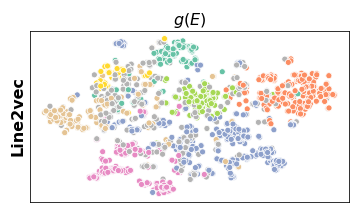}
  \includegraphics[width=\textwidth]{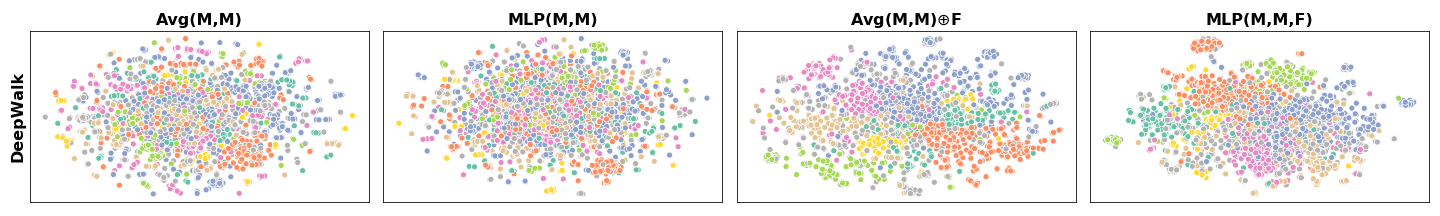}
  \includegraphics[width=\textwidth]{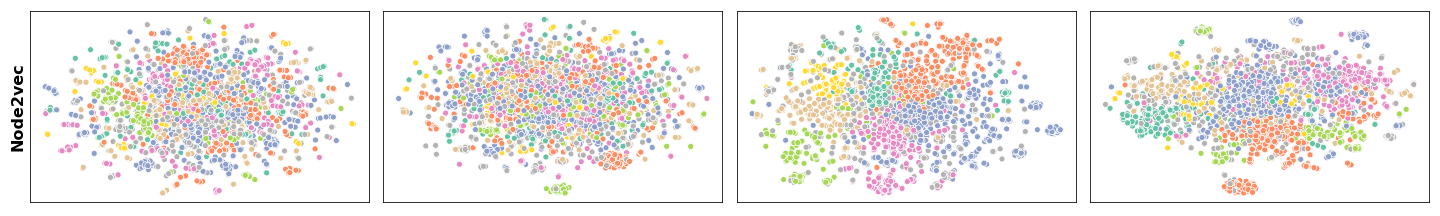}
  \includegraphics[width=\textwidth]{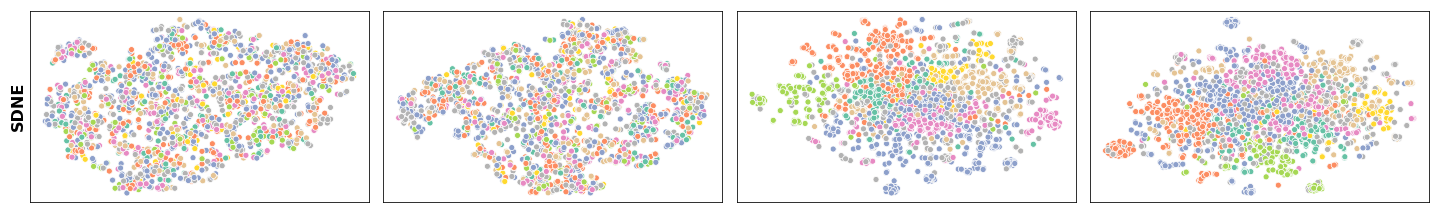}
  \includegraphics[width=\textwidth]{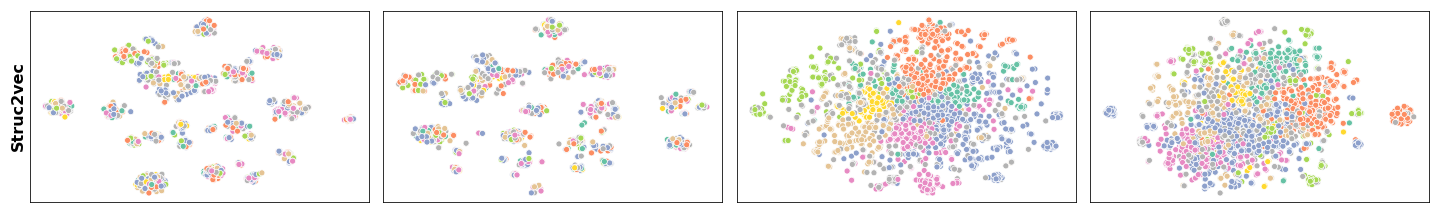}
  \includegraphics[width=\textwidth]{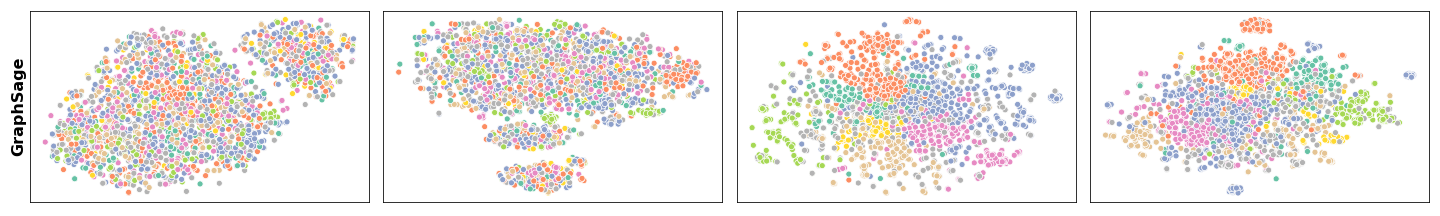}
  \includegraphics[width=\textwidth]{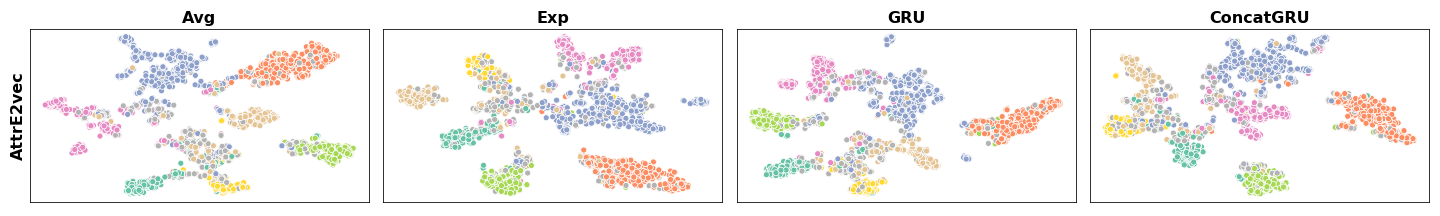}
  \caption{2-D T-SNE projections of embedding vectors for all evaluated methods. Columns denotes aggregation approach, beside $F$ that denotes the edge attributes and $g(E)$ that is an edge embedding obtained with graph structure only. Rows gather particular methods.}
  \label{fig:vectors-plots-tsne}
\end{figure*}

\subsection{Edge clustering}
Similarly to Line2vec \citep{Bandyopadhyay}, we apply the K-Means++ algorithm on the resulting embedding vectors and compute an unsupervised clustering accuracy \citep{pmlr-v48-xieb16}. We summarize the results in Table \ref{tab:edge-clustering-metrics}. Our model performs the best in all but one case and achieves significantly better results than other baseline methods. The only exception is for the Pubmed dataset, where Line2vec achieves the best clustering accuracy. Other baseline methods perform similarly as in the edge classification task. Hence, we will not discuss the details, and we encourage the reader to go through the detailed results. 

\begin{table*}[ht]
    \caption{Accuracy on edge clustering. $\mathcal{F}$ denotes the edge attributes (also referred to as "Doc2vec"), $\mathcal{M}$ -- node attributes (e.g., embeddings computed using "Node2vec"), $\oplus$ -- concatenation operator, \textbf{Avg($\mathcal{M}$,$\mathcal{M}$)} -- average operator on node embeddings, \textbf{MLP($\cdot$)} -- encoder output of MLP autoencoder trained on given attributes. \textbf{AUC} in bold shows the highest value and \textit{AUC} in italic --- the second highest value.}
    \label{tab:edge-clustering-metrics}
    \centering
    \scalebox{0.85}{
        
\begin{tabular}{|l|l|l|c|ccc|}
\hline
& \multicolumn{2}{c|}{\multirow{2}{*}{\textbf{Method group/name}}} & \textbf{Vector} & \multicolumn{3}{c|}{\textbf{Accuracy}}  \\
& \multicolumn{2}{c|}{$\;$}  & \textbf{size}  & \textbf{Citeseer} & \textbf{Cora} & \textbf{Pubmed} \\

\hline

\parbox[t]{2mm}{\multirow{18}{*}{\rotatebox[origin=c]{90}{\textbf{Transductive}}}} & \multicolumn{2}{l|}{\textbf{Edge features only; $\mathcal{F}$} (Doc2vec)} & 260 &  54.13  $\pm$ 2.73 &  54.64  $\pm$ 5.86 &  46.33  $\pm$ 1.53 \\
\cline{2-7}

& \multicolumn{2}{l|}{\textbf{Line2vec}} & 64 &  54.73  $\pm$ 2.56 &   63.50  $\pm$ 1.92 &  $\mathbf{55.26  \pm 1.36}$ \\
\cline{2-7}

& \multirow{4}{*}{\textbf{Avg($\mathcal{M}$,$\mathcal{M}$)}} & DeepWalk & 64 &  28.89  $\pm$ 1.06 &  21.93  $\pm$ 0.86 &  27.24  $\pm$ 0.50 \\
& & Node2vec      & 64 &  26.82  $\pm$ 0.67 &  21.32  $\pm$ 0.62 &  27.17  $\pm$ 0.74 \\
& & SDNE          & 64 &  21.01  $\pm$ 0.50 &  17.97  $\pm$ 0.47 &  31.38  $\pm$ 0.69 \\
& & Struc2vec     & 64 &  25.21  $\pm$ 1.33 &  20.15  $\pm$ 0.64 &  32.02  $\pm$ 1.49 \\
\cline{2-7}

& \multirow{4}{*}{\textbf{MLP($\mathcal{M}$,$\mathcal{M}$)}} & DeepWalk & 64 &  26.36 $\pm$ 1.37 &  21.06 $\pm$ 0.57 &  27.40 $\pm$ 0.93 \\
& & Node2vec & 64 &  26.37 $\pm$ 1.64 &  21.31 $\pm$ 0.98 &  27.67 $\pm$ 0.78 \\
& & SDNE & 64 &  22.27 $\pm$ 0.76 &  17.15 $\pm$ 0.36 &  28.44 $\pm$ 1.21 \\
& & Struc2vec & 64 &  24.22 $\pm$ 0.83 &  19.56 $\pm$ 0.49 &  31.31 $\pm$ 1.70 \\
\cline{2-7}

& \multirow{4}{*}{\textbf{Avg($\mathcal{M}$,$\mathcal{M}$)}$\oplus\mathcal{F}$} & DeepWalk  & 324 &  54.13  $\pm$ 2.73 &  54.70  $\pm$ 5.85 &  46.33  $\pm$ 1.53 \\
& & Node2vec  & 324 &  54.13  $\pm$ 2.73 &  54.70  $\pm$ 5.85 &  46.33  $\pm$ 1.53 \\
& & SDNE      & 324 &  55.29  $\pm$ 2.06 &  55.43  $\pm$ 4.63 &  46.33  $\pm$ 1.53 \\
& & Struc2vec & 324 &  55.59  $\pm$ 1.51 &  52.47  $\pm$ 6.52 &  46.32  $\pm$ 1.29 \\
\cline{2-7}

& \multirow{4}{*}{\textbf{MLP($\mathcal{M}$,$\mathcal{M}$,$\mathcal{F}$)}} & DeepWalk & 64 &  48.74 $\pm$ 4.03 &  47.38 $\pm$ 4.72 &  46.49 $\pm$ 1.20 \\
& & Node2vec & 64 &  50.80 $\pm$ 2.30 &  48.48 $\pm$ 3.38 &  46.15 $\pm$ 1.43 \\
& & SDNE & 64 &  46.17 $\pm$ 3.15 &  44.87 $\pm$ 3.54 &  45.74 $\pm$ 1.89 \\
& & Struc2vec & 64 &  47.35 $\pm$ 3.73 &  44.38 $\pm$ 3.04 &  45.40 $\pm$ 1.72 \\

\hline\hline

\parbox[t]{2mm}{\multirow{8}{*}{\rotatebox[origin=c]{90}{\textbf{Inductive}}}} & \textbf{Avg($\mathcal{M}$,$\mathcal{M}$)} & GraphSage & 64 & 18.79 $\pm$ 0.62 &	17.70 $\pm$ 1.05 &	27.04 $\pm$ 0.71 \\
\cline{2-7}

& \textbf{MLP($\mathcal{M}$,$\mathcal{M}$)} & GraphSage & 64 &  18.92 $\pm$ 0.98 &  17.89 $\pm$ 0.85 &  27.09 $\pm$ 0.81 \\
\cline{2-7}

& \textbf{Avg($\mathcal{M}$,$\mathcal{M}$)$\oplus\mathcal{F}$} & GraphSage & 324 &  54.06 $\pm$ 2.54 &	54.82 $\pm$ 6.86 & 46.49 $\pm$ 1.64\\
\cline{2-7}

& \textbf{MLP($\mathcal{M}$,$\mathcal{M}$,$\mathcal{F}$)} & GraphSage & 64 &  48.79 $\pm$ 4.04 &  47.49 $\pm$ 5.41 &  45.15 $\pm$ 1.54 \\
\cline{2-7}

 & \multirow{4}{*}{\textbf{AttrE2vec} (our)}                                  & Avg         & 64 &  59.82  $\pm$ 3.30 &  65.42  $\pm$ 1.71 &  48.86  $\pm$ 2.46 \\
& & Exp         & 64 &  59.07  $\pm$ 4.65 &  $\mathbf{66.36  \pm 3.62}$ &  48.02  $\pm$ 2.55 \\
& & GRU         & 64 &  $\mathit{60.16  \pm 2.25}$ &  $\mathit{66.15  \pm 3.71}$ & 49.41  $\pm$ 1.49 \\
& & ConcatGRU   & 64 &  $\mathbf{60.71  \pm 2.75}$ &  66.00  $\pm$ 2.21 &  $\mathit{50.27  \pm 3.75}$ \\
\hline
\end{tabular}
    }
\end{table*}

\subsection{Embedding visualization}
For all tested baseline methods and our proposed \texttt{AttrE2vec} method, we compute 2-dimensional projections of the produced embeddings using T-SNE \citep{t-sne} method. We visualize them in Figure \ref{fig:vectors-plots-tsne}. In our subjective opinion, these plots correspond to the AUC scores reported in Table \ref{tab:edge-clf-metrics}—the higher the AUC, the better the group separation. In details, for Doc2vec raw edge features seem to form groups, but unfortunately overlap to some degree. We cannot observe any pattern in the node embedding-based settings (\textbf{Avg($\mathcal{M}$,$\mathcal{M}$)} and \textbf{MLP($\mathcal{M}$,$\mathcal{M}$)}), they tempt to be quasi-random. When concatenated with the edge attributes (\textbf{Avg($\mathcal{M}$,$\mathcal{M}$)$\oplus\mathcal{F}$} and \textbf{MLP($\mathcal{M}$,$\mathcal{M}$,$\mathcal{F}$)}) we observe a slightly better grouping, but yet non satisfying. \texttt{AttrE2vec} model produces much more formed groups, with only a little overlapping. To summarize, based on the observed groups' separability and AUC metrics, our approach works the best among all methods.

\section{Hyperparameter Sensitivity of AttrE2vec}

\begin{figure*}
  \centering
  \includegraphics[width=0.5\textwidth]{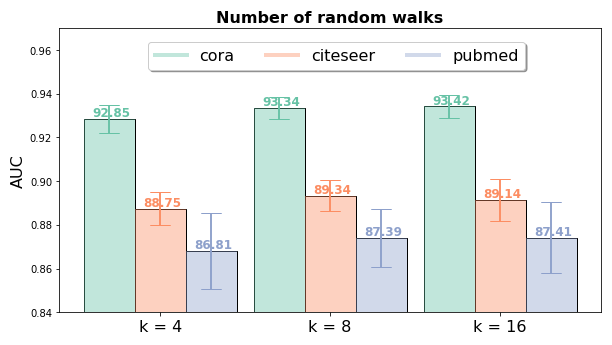}%
  \includegraphics[width=0.5\textwidth]{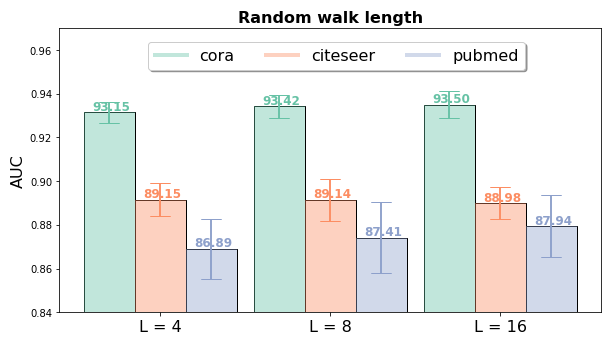}
  
  \includegraphics[width=0.5\textwidth]{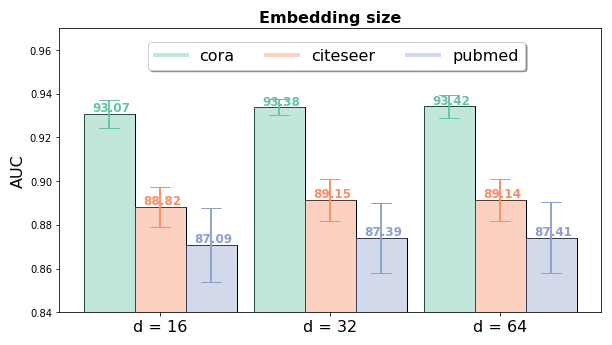}%
  \includegraphics[width=0.5\textwidth]{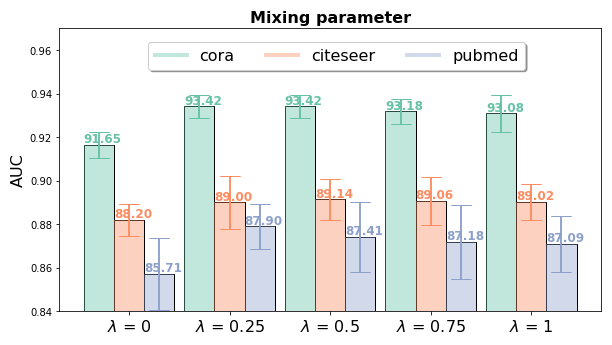}

  \caption{Effects of hyperparameters on Cora, Citeseer and Pubmed datasets.}
  \label{fig:hps}
\end{figure*}
We investigate hyperparameters' effect considering each of them independently, i.e., setting a given parameter and preserving default values for all other parameters. The evaluation is applied for our model's two inductive variants: with the Average aggregator and with the GRU aggregator. We use all three datasets (Cora, Citeseer, Pubmed) and report the AUC values. We choose following hyperparameter value sets (values with an asterisk denote the default value for that parameter):
\begin{itemize}
    \item length of random walk: $L = \{4, \mathbf{8^*}, 16\}$,
    \item number of random walks: $k = \{4, 8, \mathbf{16^*}\}$,
    \item embedding size: $d = \{16, 32, \mathbf{64^*}\}$,
    \item mixing parameter: $\lambda = \{0, 0.25, \mathbf{0.5^*}, 0.75, 1\}$.
\end{itemize}
The results of all experiments are summarized in Figure \ref{fig:hps}. We observe that for both aggregation variants, Avg and GRU, the trends are similar, so we will include and discuss them based only on the Average aggregator. 

In general, the higher the number of random walks $k$ and the length of a single random walk $L$, the better results are achieved. One may require higher values of these parameters, but it significantly increases the random walk computation time and the model training itself.

Unsurprisingly, the embedding size (embedding dimension) also follows the same trend. With more dimensions, we can fit more information into the created representations. However, as an embedding goal is to find \textbf{low-dimensional} vector representations, we should keep reasonable dimensionality. Our chosen values (16, 32, 64) seem plausible while working with 260-dimensional edge features.

\begin{figure}
    \centering
    \includegraphics[width=0.7\columnwidth]{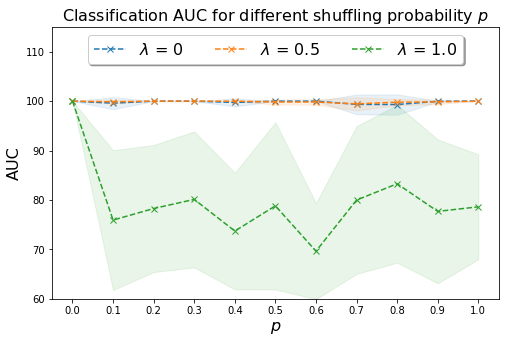}
    \caption{\texttt{AttrE2vec} performance for various noise levels $p$ and mixing parameter values $\lambda \in \{0, 0.5, 1\}$.}
    \label{fig:ablation-plot}
\end{figure}

As for loss mixing parameter $\lambda$, we observe that too high values negatively influence the model performance. The greater the value, the more critical the structural loss becomes. Simultaneously the feature loss becomes less relevant. Choosing $\lambda = 0$ causes the loss function to consider feature reconstruction only and completely ignores the embedding loss. This yields significantly worse results and confirms that our approach of combining both feature reconstruction and structural embedding loss is justified. In general, the best values are achieved for setting an equal influence of both loss factors ($\lambda = 0.5$).

\section{Ablation study}
\begin{figure*}
    \centering
    \includegraphics[width=\textwidth]{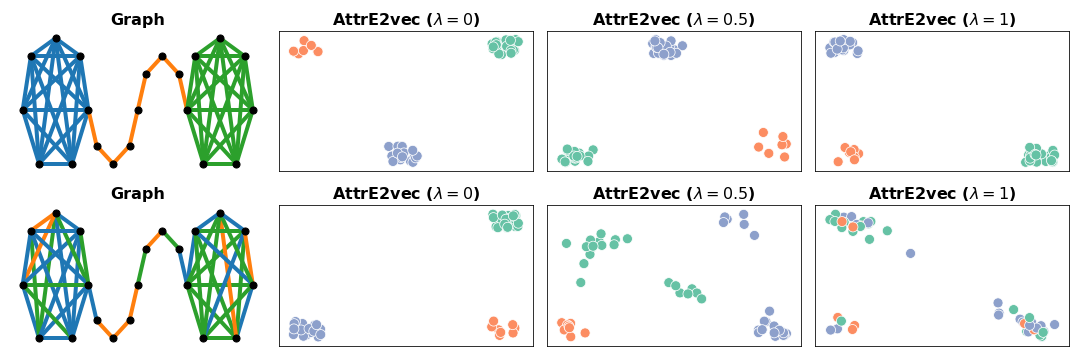}
    \caption{2-D representations of ideal and noisy graph edges using \texttt{AttrE2vec} with $\lambda \in \{0, 0.5, 1\}$.\label{fig:ablation-full}}
\end{figure*}
We performed an ablation study to check whether our method \texttt{AttrE2vec} is invariant to introduced noise in an artificially generated network. We use a barbell graph, which consists of two fully connected graphs and a path which connects them (see: Figure \ref{fig:intro-figure}). The graph has seven nodes in each full graph and seven nodes in the path -- a total of 50 edges. Next, we generate features from 3 clusters in a 200-dimensional space using isotropic Gaussian blobs. We assign the features to 3 parts of the graph: the first to the edges in one of the full graphs, the second to the edges in the path and the third to the edges in the other full graph. The edge classes are matching the feature clusters (i.e., three classes). Therefore, the structure is aligned with the features, so any good structure based embedding method can fit this data very well (see: Figure \ref{fig:intro-figure}). A problem occurs when the features (and hence the classes) are shuffled within the graph structure. Methods that employ only a structural loss function will fail. We want to check how our model \texttt{AttrE2vec}, which includes both structural and feature-based loss, performs with different amount of such noise.

We will use the graph mentioned above and introduce noise by shuffling $p\%$ of all edge pairs, which are from different classes, i.e., an edge with class 2 (originally located in the path) may be swapped with one from the full graphs (classes 1 or 3). We use our AttrE2vec model with an Average aggregator in the transductive setting (due to the graph size) and report the edge classification AUC for different values of $p \in \{0, 0.1, \ldots, 0.5, \ldots, 0.9, 1\}$ and $\lambda \in \{0, 0.5, 1\}$. The values of the mixing parameter $\lambda$ allow us to check how the model behaves when working only with a feature-based loss ($\lambda = 0$), only with a structural loss ($\lambda = 1$), and with both losses at equal importance ($\lambda = 0.5$). We train our model for five epochs and repeat the computations ten times for every $(p, \lambda)$ pair, due to the shuffling procedure's randomness. We report the mean and standard deviation of the AUC value in Figure \ref{fig:ablation-plot}.

Using only the feature loss or a combination of both losses allows us to achieve nearly 100\% AUC in the classification task. The fluctuations appear due to the low number of training epochs and the local optima problem. The performance of the model that uses only structural loss ($\lambda = 1$) decreases with higher shuffling probabilities, and from a certain point, it starts improving slightly because shuffling results in a complete swap of two classes, i.e., all features and classes from one graph part are exchanged with all features and classes from another part of the graph. 

We also demonstrate how our method reacts on noisy data with various $\lambda \in \{0, 0.5, 1\}$. There are two graphs: one where the features are aligned to substructures of the graph and the second with shuffled features (ca. 50\%), see Figure \ref{fig:ablation-full}. Keeping \texttt{AttrE2vec} with $\lambda = 0.5$ allows to represent noisy graphs fairly.

\section{Conclusions and future work}

We introduce \texttt{AttrE2vec} – the novel unsupervised and inductive embedding model to learn attributed edge embeddings by leveraging on the self-attention network with auto-encoder over attribute space and structural loss on aggregated random walks. \textit{Attre2vec} can directly aggregate feature information from edges and nodes at many hops away to infer embeddings not only for present nodes, but also for new nodes. Extensive experimental results show that \texttt{AttrE2vec} obtains the state-of-the-art results in edge classification and clustering on CORA, PUBMED and CITESEER.

\section*{Acknowledgments}
The work was partially supported by the National Science Centre, Poland grant No. 2016/21/D/ST6/02948, and 2016/23/B/ST6/01735, as well as by the Department of Computational Intelligence, Wrocław University of Science and Technology statutory funds.

\bibliographystyle{elsarticle-num}
\bibliography{references-tk}

\begin{thebibliography}{10}
\expandafter\ifx\csname url\endcsname\relax
  \def\url#1{\texttt{#1}}\fi
\expandafter\ifx\csname urlprefix\endcsname\relax\def\urlprefix{URL }\fi
\expandafter\ifx\csname href\endcsname\relax
  \def\href#1#2{#2} \def\path#1{#1}\fi

\bibitem{Hua}
W.~Hu, M.~Fey, M.~Zitnik, Y.~Dong, H.~Ren, B.~Liu, M.~Catasta, J.~Leskovec,
  R.~Barzilay, P.~Battaglia, Y.~Bengio, M.~Bronstein, S.~G{\"{u}}nnemann,
  W.~Hamilton, T.~Jaakkola, S.~Jegelka, M.~Nickel, C.~Re, L.~Song, J.~Tang,
  M.~Welling, R.~Zemel, \href{http://arxiv.org/abs/2005.00687}{{Open graph
  benchmark: Datasets for machine learning on graphs}} (may 2020).
\newblock \href {http://arxiv.org/abs/2005.00687} {\path{arXiv:2005.00687}}.
\newline\urlprefix\url{http://arxiv.org/abs/2005.00687}

\bibitem{Zhang2018a}
D.~Zhang, J.~Yin, X.~Zhu, C.~Zhang, {Network Representation Learning: A
  Survey}, IEEE Transactions on Big Data 6~(1) (2018) 3--28.
\newblock \href {http://dx.doi.org/10.1109/tbdata.2018.2850013}
  {\path{doi:10.1109/tbdata.2018.2850013}}.

\bibitem{Wu2019}
Z.~Wu, S.~Pan, F.~Chen, G.~Long, C.~Zhang, P.~S. Yu, {A Comprehensive Survey on
  Graph Neural Networks}, IEEE Transactions on Neural Networks and Learning
  Systems (2019) 1--21\href {http://dx.doi.org/10.1109/TNNLS.2020.2978386}
  {\path{doi:10.1109/TNNLS.2020.2978386}}.

\bibitem{Lia}
B.~Li, D.~Pi, {Network representation learning: a systematic literature
  review}, Neural Computing and Applications 32~(21) (2020) 16647--16679.
\newblock \href {http://dx.doi.org/10.1007/s00521-020-04908-5}
  {\path{doi:10.1007/s00521-020-04908-5}}.

\bibitem{Chami2020}
I.~Chami, S.~Abu-El-Haija, B.~Perozzi, C.~R{\'{e}}, K.~Murphy,
  \href{http://arxiv.org/abs/2005.03675}{{Machine Learning on Graphs: A Model
  and Comprehensive Taxonomy}} (2020).
\newline\urlprefix\url{http://arxiv.org/abs/2005.03675}

\bibitem{autoweight}
S.~Bahrami, F.~Dornaika, A.~Bosaghzadeh, \href{www.scopus.com}{Joint
  auto-weighted graph fusion and scalable semi-supervised learning},
  Information Fusion 66 (2021) 213--228.
\newline\urlprefix\url{www.scopus.com}

\bibitem{Grover2016}
A.~Grover, J.~Leskovec, {Node2vec: Scalable feature learning for networks}, in:
  Proceedings of the ACM SIGKDD International Conference on Knowledge Discovery
  and Data Mining, Vol. 13-17-Augu, 2016, pp. 855--864.
\newblock \href {http://dx.doi.org/10.1145/2939672.2939754}
  {\path{doi:10.1145/2939672.2939754}}.

\bibitem{Perozzia}
B.~Perozzi, R.~Al-Rfou, S.~Skiena,
  \href{http://dl.acm.org/citation.cfm?doid=2623330.2623732}{{DeepWalk: Online
  Learning of Social Representations Bryan}}, in: Proceedings of the 20th ACM
  SIGKDD international conference on Knowledge discovery and data mining - KDD
  '14, ACM Press, New York, New York, USA, 2014, pp. 701--710.
\newblock \href {http://dx.doi.org/10.1145/2623330.2623732}
  {\path{doi:10.1145/2623330.2623732}}.
\newline\urlprefix\url{http://dl.acm.org/citation.cfm?doid=2623330.2623732}

\bibitem{Kipf2019}
T.~N. Kipf, M.~Welling, \href{http://arxiv.org/abs/1609.02907}{{Semi-supervised
  classification with graph convolutional networks}}, in: 5th International
  Conference on Learning Representations, ICLR 2017 - Conference Track
  Proceedings, International Conference on Learning Representations, ICLR,
  2017, pp. 1--14.
\newblock \href {http://arxiv.org/abs/1609.02907} {\path{arXiv:1609.02907}}.
\newline\urlprefix\url{http://arxiv.org/abs/1609.02907}

\bibitem{DongMetapath2vec:Networks}
Y.~Dong, N.~V. Chawla, A.~Swami,
  \href{https://dl.acm.org/doi/10.1145/3097983.3098036}{{Metapath2vec: Scalable
  representation learning for heterogeneous networks}}, in: Proceedings of the
  ACM SIGKDD International Conference on Knowledge Discovery and Data Mining,
  Vol. Part F1296, ACM, New York, NY, USA, 2017, pp. 135--144.
\newblock \href {http://dx.doi.org/10.1145/3097983.3098036}
  {\path{doi:10.1145/3097983.3098036}}.
\newline\urlprefix\url{https://dl.acm.org/doi/10.1145/3097983.3098036}

\bibitem{covid19gcn}
S.~. Wang, V.~V. Govindaraj, J.~M. Górriz, X.~Zhang, Y.~. Zhang,
  \href{www.scopus.com}{Covid-19 classification by fgcnet with deep feature
  fusion from graph convolutional network and convolutional neural network},
  Information Fusion 67 (2021) 208--229, cited By :1.
\newline\urlprefix\url{www.scopus.com}

\bibitem{Garcia-Duran2017}
A.~Garc{\'{i}}a-Dur{\'{a}}n, M.~Niepert, {Learning graph representations with
  embedding propagation}, in: Advances in Neural Information Processing
  Systems, Vol. 2017-Decem, 2017, pp. 5120--5131.

\bibitem{Hamiltona}
W.~L. Hamilton, R.~Ying, J.~Leskovec, {Inductive representation learning on
  large graphs}, in: Advances in Neural Information Processing Systems, Vol.
  2017-Decem, 2017, pp. 1025--1035.

\bibitem{Velickovic2018}
P.~Veli{\v{c}}kovi{\'{c}}, A.~Casanova, P.~Li{\`{o}}, G.~Cucurull, A.~Romero,
  Y.~Bengio, {Graph attention networks}, in: 6th International Conference on
  Learning Representations, ICLR 2018 - Conference Track Proceedings,
  International Conference on Learning Representations, ICLR, 2018, pp. 1--12.
\newblock \href {http://arxiv.org/abs/1710.10903} {\path{arXiv:1710.10903}}.

\bibitem{Wanga}
D.~Wang, P.~Cui, W.~Zhu, {Structural deep network embedding}, in: Proceedings
  of the ACM SIGKDD International Conference on Knowledge Discovery and Data
  Mining, Vol. 13-17-Augu, 2016, pp. 1225--1234.
\newblock \href {http://dx.doi.org/10.1145/2939672.2939753}
  {\path{doi:10.1145/2939672.2939753}}.

\bibitem{Yang2015}
C.~Yang, Z.~Liu, D.~Zhao, M.~Sun, E.~Y. Chang, {Network representation learning
  with rich text information}, in: IJCAI International Joint Conference on
  Artificial Intelligence, Vol. 2015-Janua, 2015, pp. 2111--2117.

\bibitem{ahng}
M.~Liu, J.~Liu, Y.~Chen, M.~Wang, H.~Chen, Q.~Zheng,
  \href{www.scopus.com}{Ahng: Representation learning on attributed
  heterogeneous network}, Information Fusion 50 (2019) 221--230, cited By :3.
\newline\urlprefix\url{www.scopus.com}

\bibitem{Lan2020ImprovingContent}
L.~Lan, P.~Wang, J.~Zhao, J.~Tao, J.~Lui, X.~Guan, {Improving network embedding
  with partially available vertex and edge content}, Information Sciences 512
  (2020) 935--951.
\newblock \href {http://dx.doi.org/10.1016/j.ins.2019.09.083}
  {\path{doi:10.1016/j.ins.2019.09.083}}.

\bibitem{Li2020Multi-sourceEmbedding}
B.~Li, D.~Pi, Y.~Lin, I.~Khan, L.~Cui, {Multi-source information fusion based
  heterogeneous network embedding}, Information Sciences 534 (2020) 53--71.
\newblock \href {http://dx.doi.org/10.1016/j.ins.2020.05.012}
  {\path{doi:10.1016/j.ins.2020.05.012}}.

\bibitem{ZhangHeterogeneousNetwork}
C.~Zhang, D.~Song, C.~Huang, A.~Swami, N.~V. Chawla,
  \href{https://dl.acm.org/doi/10.1145/3292500.3330961}{{Heterogeneous graph
  neural network}}, in: Proceedings of the ACM SIGKDD International Conference
  on Knowledge Discovery and Data Mining, ACM, New York, NY, USA, 2019, pp.
  793--803.
\newblock \href {http://dx.doi.org/10.1145/3292500.3330961}
  {\path{doi:10.1145/3292500.3330961}}.
\newline\urlprefix\url{https://dl.acm.org/doi/10.1145/3292500.3330961}

\bibitem{Gao2018a}
H.~Gao, H.~Huang, {Deep attributed network embedding}, in: IJCAI International
  Joint Conference on Artificial Intelligence, Vol. 2018-July, 2018, pp.
  3364--3370.
\newblock \href {http://dx.doi.org/10.24963/ijcai.2018/467}
  {\path{doi:10.24963/ijcai.2018/467}}.

\bibitem{Bandyopadhyay}
S.~Bandyopadhyay, A.~Biswas, N.~Murty, R.~Narayanam, {Beyond node embedding: A
  direct unsupervised edge representation framework for homogeneous networks}
  (2019).
\newblock \href {http://arxiv.org/abs/1912.05140} {\path{arXiv:1912.05140}}.

\bibitem{Chen2020RelationEmbedding}
Y.~Chen, T.~Qian, {Relation constrained attributed network embedding},
  Information Sciences 515 (2020) 341--351.
\newblock \href {http://dx.doi.org/10.1016/j.ins.2019.12.033}
  {\path{doi:10.1016/j.ins.2019.12.033}}.

\bibitem{Bandyopadhyay2018}
S.~Bandyopadhyay, H.~Kara, A.~Kannan, M.~N. Murty, {FSCNMF: Fusing structure
  and content via non-negative matrix factorization for embedding information
  networks} (2018).
\newblock \href {http://arxiv.org/abs/1804.05313} {\path{arXiv:1804.05313}}.

\bibitem{Nozza2020CAGE:Embedding}
D.~Nozza, E.~Fersini, E.~Messina, {CAGE: Constrained deep Attributed Graph
  Embedding}, Information Sciences 518 (2020) 56--70.
\newblock \href {http://dx.doi.org/10.1016/j.ins.2019.12.082}
  {\path{doi:10.1016/j.ins.2019.12.082}}.

\bibitem{Kim}
J.~Kim, T.~Kim, S.~Kim, C.~D. Yoo, {Edge-labeling graph neural network for
  few-shot learning}, in: Proceedings of the IEEE Computer Society Conference
  on Computer Vision and Pattern Recognition, Vol. 2019-June, 2019, pp. 11--20.
\newblock \href {http://arxiv.org/abs/1905.01436} {\path{arXiv:1905.01436}},
  \href {http://dx.doi.org/10.1109/CVPR.2019.00010}
  {\path{doi:10.1109/CVPR.2019.00010}}.

\bibitem{Li2019}
Q.~Li, Z.~Cao, J.~Zhong, Q.~Li, {Graph representation learning with encoding
  edges}, Neurocomputing 361 (2019) 29--39.
\newblock \href {http://dx.doi.org/10.1016/j.neucom.2019.07.076}
  {\path{doi:10.1016/j.neucom.2019.07.076}}.

\bibitem{Gong2019}
L.~Gong, Q.~Cheng, {Exploiting edge features for graph neural networks}, in:
  Proceedings of the IEEE Computer Society Conference on Computer Vision and
  Pattern Recognition, 2019, pp. 9203--9211.
\newblock \href {http://dx.doi.org/10.1109/CVPR.2019.00943}
  {\path{doi:10.1109/CVPR.2019.00943}}.

\bibitem{Aggarwal2016a}
C.~Aggarwal, G.~He, P.~Zhao, {Edge classification in networks}, in: 2016 IEEE
  32nd International Conference on Data Engineering, ICDE 2016, Institute of
  Electrical and Electronics Engineers Inc., 2016, pp. 1038--1049.
\newblock \href {http://dx.doi.org/10.1109/ICDE.2016.7498311}
  {\path{doi:10.1109/ICDE.2016.7498311}}.

\bibitem{Simonovsky}
M.~Simonovsky, N.~Komodakis, {Dynamic edge-conditioned filters in convolutional
  neural networks on graphs}, in: Proceedings - 30th IEEE Conference on
  Computer Vision and Pattern Recognition, CVPR 2017, Vol. 2017-Janua, 2017,
  pp. 29--38.
\newblock \href {http://dx.doi.org/10.1109/CVPR.2017.11}
  {\path{doi:10.1109/CVPR.2017.11}}.

\bibitem{Bui2018}
T.~D. Bui, S.~Ravi, V.~Ramavajjala, {Neural Graph Learning: Training Neural
  Networks Using Graphs}, dl.acm.org 2018-Febua (2018) 64--71.
\newblock \href {http://dx.doi.org/10.1145/3159652.3159731}
  {\path{doi:10.1145/3159652.3159731}}.

\bibitem{Wang2019a}
Y.~Wang, Y.~Sun, M.~M. Bronstein, J.~M. Solomon, Z.~Liu, S.~E. Sarma, {Dynamic
  Graph CNN for Learning on Point Clouds}, ACM Transactions on Graphics 38~(5)
  (2019) 146.
\newblock \href {http://dx.doi.org/10.1145/3326362}
  {\path{doi:10.1145/3326362}}.

\bibitem{Wanyan2020}
T.~Wanyan, C.~Zhang, A.~Azad, X.~Liang, D.~Li, Y.~Ding,
  \href{http://arxiv.org/abs/2004.01375}{{Attribute2vec: Deep network embedding
  through multi-filtering GCN}} (apr 2020).
\newblock \href {http://arxiv.org/abs/2004.01375} {\path{arXiv:2004.01375}}.
\newline\urlprefix\url{http://arxiv.org/abs/2004.01375}

\bibitem{Tang2015}
J.~Tang, M.~Qu, M.~Wang, M.~Zhang, J.~Yan, Q.~Mei, {LINE: Large-scale
  information network embedding}, in: WWW 2015 - Proceedings of the 24th
  International Conference on World Wide Web, 2015, pp. 1067--1077.
\newblock \href {http://dx.doi.org/10.1145/2736277.2741093}
  {\path{doi:10.1145/2736277.2741093}}.

\bibitem{Ribeiro2017}
L.~F. Ribeiro, P.~H. Saverese, D.~R. Figueiredo, {Struc2vec: Learning node
  representations from structural identity}, in: Proceedings of the ACM SIGKDD
  International Conference on Knowledge Discovery and Data Mining, Vol. Part
  F1296, 2017, pp. 385--394.
\newblock \href {http://dx.doi.org/10.1145/3097983.3098061}
  {\path{doi:10.1145/3097983.3098061}}.

\bibitem{node2vec}
A.~Grover, J.~Leskovec, node2vec: Scalable feature learning for networks, in:
  Proceedings of the 22nd ACM SIGKDD international conference on Knowledge
  discovery and data mining, ACM, 2016, pp. 855--864.

\bibitem{Chung2014EmpiricalModeling}
J.~Chung, C.~Gulcehre, K.~Cho, Y.~Bengio,
  \href{http://arxiv.org/abs/1412.3555}{{Empirical Evaluation of Gated
  Recurrent Neural Networks on Sequence Modeling}} (dec 2014).
\newblock \href {http://arxiv.org/abs/1412.3555} {\path{arXiv:1412.3555}}.
\newline\urlprefix\url{http://arxiv.org/abs/1412.3555}

\bibitem{dvc}
R.~Kuprieiev, D.~Petrov, R.~Valles, P.~Redzyński, C.~da~Costa-Luis,
  A.~Schepanovski, I.~Shcheklein, S.~Pachhai, J.~Orpinel, F.~Santos, A.~Sharma,
  Zhanibek, D.~Hodovic, P.~Rowlands, Earl, A.~Grigorev, N.~Dash, G.~Vyshnya,
  maykulkarni, Vera, M.~Hora, xliiv, W.~Baranowski, S.~Mangal, C.~Wolff,
  nik123, O.~Yoktan, K.~Benoy, A.~Khamutov, A.~Maslakov,
  \href{https://doi.org/10.5281/zenodo.3859749}{Dvc: Data version control - git
  for data \& models} (May 2020).
\newblock \href {http://dx.doi.org/10.5281/zenodo.3859749}
  {\path{doi:10.5281/zenodo.3859749}}.
\newline\urlprefix\url{https://doi.org/10.5281/zenodo.3859749}

\bibitem{cora_dataset}
P.~Sen, G.~Namata, M.~Bilgic, L.~Getoor, B.~Galligher, T.~Eliassi-Rad,
  \href{https://ojs.aaai.org/index.php/aimagazine/article/view/2157}{Collective
  classification in network data}, AI Magazine 29~(3) (2008) 93.
\newblock \href {http://dx.doi.org/10.1609/aimag.v29i3.2157}
  {\path{doi:10.1609/aimag.v29i3.2157}}.
\newline\urlprefix\url{https://ojs.aaai.org/index.php/aimagazine/article/view/2157}

\bibitem{pubmed_dataset}
G.~Namata, B.~London, L.~Getoor, B.~Huang, {Query-driven Active Surveying for
  Collective Classification}, in: Proceedings ofthe Workshop on Mining and
  Learn- ing with Graphs, Edinburgh, Scotland, UK., 2012, pp. 1--8.

\bibitem{Le}
Q.~Le, T.~Mikolov, \href{http://arxiv.org/abs/1405.4053}{{Distributed
  representations of sentences and documents}}, in: 31st International
  Conference on Machine Learning, ICML 2014, Vol.~4, 2014, pp. 2931--2939.
\newblock \href {http://arxiv.org/abs/1405.4053} {\path{arXiv:1405.4053}}.
\newline\urlprefix\url{http://arxiv.org/abs/1405.4053}

\bibitem{scikit-learn}
F.~Pedregosa, G.~Varoquaux, A.~Gramfort, V.~Michel, B.~Thirion, O.~Grisel,
  M.~Blondel, P.~Prettenhofer, R.~Weiss, V.~Dubourg, J.~Vanderplas, A.~Passos,
  D.~Cournapeau, M.~Brucher, M.~Perrot, E.~Duchesnay, Scikit-learn: Machine
  learning in {P}ython, Journal of Machine Learning Research 12 (2011)
  2825--2830.

\bibitem{sdne}
D.~Wang, P.~Cui, W.~Zhu,
  \href{http://doi.acm.org/10.1145/2939672.2939753}{Structural deep network
  embedding}, in: Proceedings of the 22Nd ACM SIGKDD International Conference
  on Knowledge Discovery and Data Mining, KDD '16, ACM, New York, NY, USA,
  2016, pp. 1225--1234.
\newblock \href {http://dx.doi.org/10.1145/2939672.2939753}
  {\path{doi:10.1145/2939672.2939753}}.
\newline\urlprefix\url{http://doi.acm.org/10.1145/2939672.2939753}

\bibitem{QuocNguyenAModel}
D.~Q. Nguyen, T.~D. Nguyen, D.~Phung, \href{http://arxiv.org/abs/2006.12100}{{A
  self-attention network based node embedding model}} (jun 2020).
\newblock \href {http://arxiv.org/abs/2006.12100} {\path{arXiv:2006.12100}}.
\newline\urlprefix\url{http://arxiv.org/abs/2006.12100}

\bibitem{Loshchilov}
I.~Loshchilov, F.~Hutter, \href{http://arxiv.org/abs/1711.05101}{{Decoupled
  Weight Decay Regularization}} (nov 2017).
\newblock \href {http://arxiv.org/abs/1711.05101} {\path{arXiv:1711.05101}}.
\newline\urlprefix\url{http://arxiv.org/abs/1711.05101}

\bibitem{pmlr-v48-xieb16}
J.~Xie, R.~Girshick, A.~Farhadi,
  \href{http://proceedings.mlr.press/v48/xieb16.html}{Unsupervised deep
  embedding for clustering analysis}, in: M.~F. Balcan, K.~Q. Weinberger
  (Eds.), Proceedings of The 33rd International Conference on Machine Learning,
  Vol.~48 of Proceedings of Machine Learning Research, PMLR, New York, New
  York, USA, 2016, pp. 478--487.
\newline\urlprefix\url{http://proceedings.mlr.press/v48/xieb16.html}

\bibitem{t-sne}
L.~van~der Maaten, G.~Hinton,
  \href{http://www.jmlr.org/papers/v9/vandermaaten08a.html}{Visualizing data
  using {t-SNE}}, Journal of Machine Learning Research 9 (2008) 2579--2605.
\newline\urlprefix\url{http://www.jmlr.org/papers/v9/vandermaaten08a.html}

\end{thebibliography}

\end{document}